\documentclass[10pt,twocolumn,letterpaper]{article}

\usepackage[]{cvpr}      

\usepackage[dvipsnames]{xcolor}

\definecolor{cvprblue}{rgb}{0.21,0.49,0.74}
\usepackage[pagebackref,breaklinks,colorlinks,citecolor=cvprblue]{hyperref}

\usepackage{dsfont}
\usepackage{graphicx}
\usepackage{amsmath}
\usepackage{amssymb}
\usepackage{multirow}
\usepackage{booktabs}
\usepackage{tabularx}
\usepackage{makecell}
\usepackage{colortbl}
\def\paperID{6944} 
\def\confName{CVPR}
\def\confYear{2024}

\title{Text as Image: Learning Transferable Adapter for Multi-Label Classification}

\author{Xuelin Zhu$^1$\quad  Jiuxin Cao$^1$\thanks{Corresponding author.}\quad  Jian liu$^2$\quad  Dongqi Tang$^2$\quad  Furong Xu$^2$\quad  Weijia Liu$^1$ \\ Jiawei Ge$^1$\quad  Bo Liu$^1$\quad  Qingpei Guo$^2$\quad  Tianyi Zhang$^2$ \\
$^1$Southeast University, \quad $^2$Ant Group, Hangzhou, China \\
{\tt\small \{zhuxuelin,jx.cao,weijia-liu,jiawei\_ge,bliu\}@seu.edu.cn, rex.lj@antgroup.com}}

\begin{document}
\maketitle

\begin{abstract}
Pre-trained vision-language models have notably accelerated progress of open-world concept recognition. Their impressive zero-shot ability has recently been transferred to multi-label image classification via prompt tuning, enabling to discover novel labels in an open-vocabulary manner. However, this paradigm suffers from non-trivial training costs, and becomes computationally prohibitive for a large number of candidate labels. To address this issue, we note that vision-language pre-training aligns images and texts in a unified embedding space, making it potential for an adapter network to identify labels in visual modality while be trained in text modality. To enhance such cross-modal transfer ability, a simple yet effective method termed random perturbation is proposed, which enables the adapter to search for potential visual embeddings by perturbing text embeddings with noise during training, resulting in better performance in visual modality. Furthermore, we introduce an effective approach to employ large language models for multi-label instruction-following text generation. In this way, a fully automated pipeline for visual label recognition is developed without relying on any manual data. Extensive experiments on public benchmarks show the superiority of our method in various multi-label classification tasks.


\end{abstract}

\section{Introduction}




Recently, vision-language (VL) pre-trained models have attracted increasing attention for their impressive zero-shot abilities across diverse downstream tasks, greatly accelerating the progress towards open-world concept recognition. Specifically, prompt learning has emerged as an dominant way for enhancing zero-shot abilities of VL models in multi-label zero (few) -shot learning. For example, CoOp \cite{zhou2022learning} models contexts of a prompt with learnable vectors. CoCoOp \cite{zhou2022conditional} extends CoOp by learning a network to yield an input-conditional vector for each image. DualCoOp \cite{sun2022dualcoop} encodes positive and negative contexts with class names as part of the prompts. TaI-DPT \cite{guo2023texts} uses both coarse-grained and ﬁne-grained features for prompt tuning.

\begin{figure}
    \centering
    \includegraphics[width=1\linewidth]{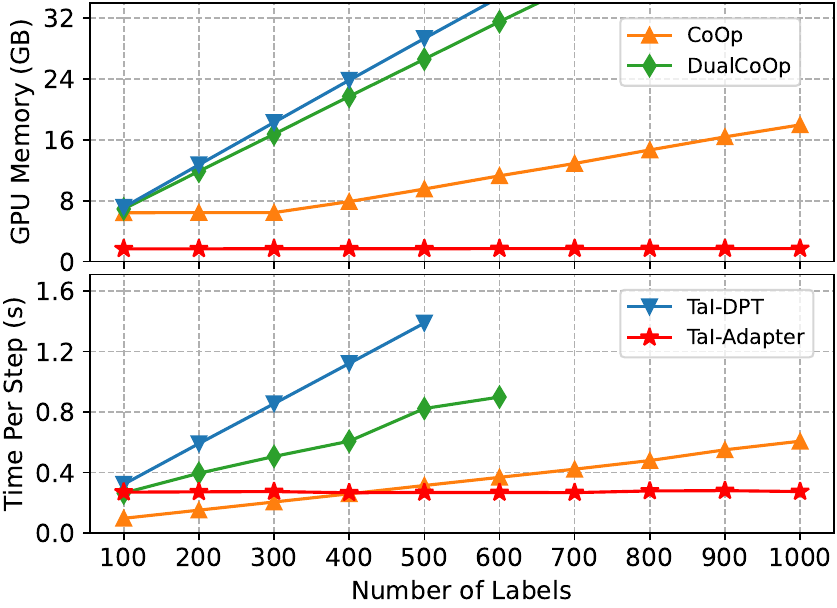}
    \caption{Comparison of the impact of label size on model training speed and memory consumption. All results are reported on a V100 GPU with a batch size of 64. Clearly, all the prompt learning based methods like CoOp \cite{zhou2022learning}, DualCoOp \cite{sun2022dualcoop} and TaI-DPT \cite{guo2023texts}, experiences substantial increases in both memory consumption and training time as the number of labels increase. In contrast, our method (TaI-Adapter) remains unaffected by label size.}
    \label{fig:memory}
\end{figure}

Although these prompt learning based approaches have achieved promising progress, they rely on the VL text encoder for encoding the prompt texts. This reliance noticeably escalates computational costs and memory consumption, exacerbating the situation particularly when label size grows larger, as depicted in Fig. \ref{fig:memory}. Specifically, taking the Open Images dataset \cite{kuznetsova2020open} as an example, it has a total of 7186 human-verified trainable labels. In the prompt learning setting, these labels are initially merged with learnable prompt contexts and subsequently fed into the VL text encoder for encoding. Indubitably, repeatedly executing this process at each training step would render these prompt learning methods computationally prohibitive.

To address above issue, we note that the primary objective of vision-language pre-training is to pull the proximity of embeddings for images and texts with similar semantics, and push embeddings with dissimilar semantics apart. Consequentially, images and texts are well aligned in the VL embedding space. Hence, it is potential to treat texts as images to train a network for visual label recognition, that is, train the network in text modality, but identify labels in visual modality. This paradigm has two benefits: 1) the computational burden of encoding prompt texts by the VL text encoder is avoided; 2) a lightweight adapter network can be learned for visual label recognition using only labeled texts, which are more easily to collect than image data.


To acquire labeled texts, some attempts \cite{zhang2023recognize, guo2023texts, huang2023tag2text} have been made to derive labels from the off-the-shelf texts, either sourced from VL datasets or obtained from human-annotated image caption datasets. However, these methods necessitate extra workloads in text pre-processing, such as noun filtering and synonym mapping. Unfortunately, this cumbersome process commonly proves to be laborious and fragile, and to inevitably bring noisy data. In light of the inherent limitations of deriving labels from texts, we resort to generate texts from labels instead. Recently, there have been several successful practices \cite{liu2023visual, zhu2023minigpt} in generating desired data by imposing instructions on large language models (LLM). This paradigm, known as instruction-following data generation, has emerged as a prospective substitute for the labor-intensive process of manually annotating data. Inspired by this, we advocate employing LLM to create multi-label instruction-following texts for labels. Specifically, we design a template that, when populated with diverse labels, serves as an instruction to drive LLM to generate relevant texts. In this way, we are able to effortlessly collect a substantial corpus of texts for any label set of interest.



\begin{figure}
    \centering
    \includegraphics[width=1\linewidth]{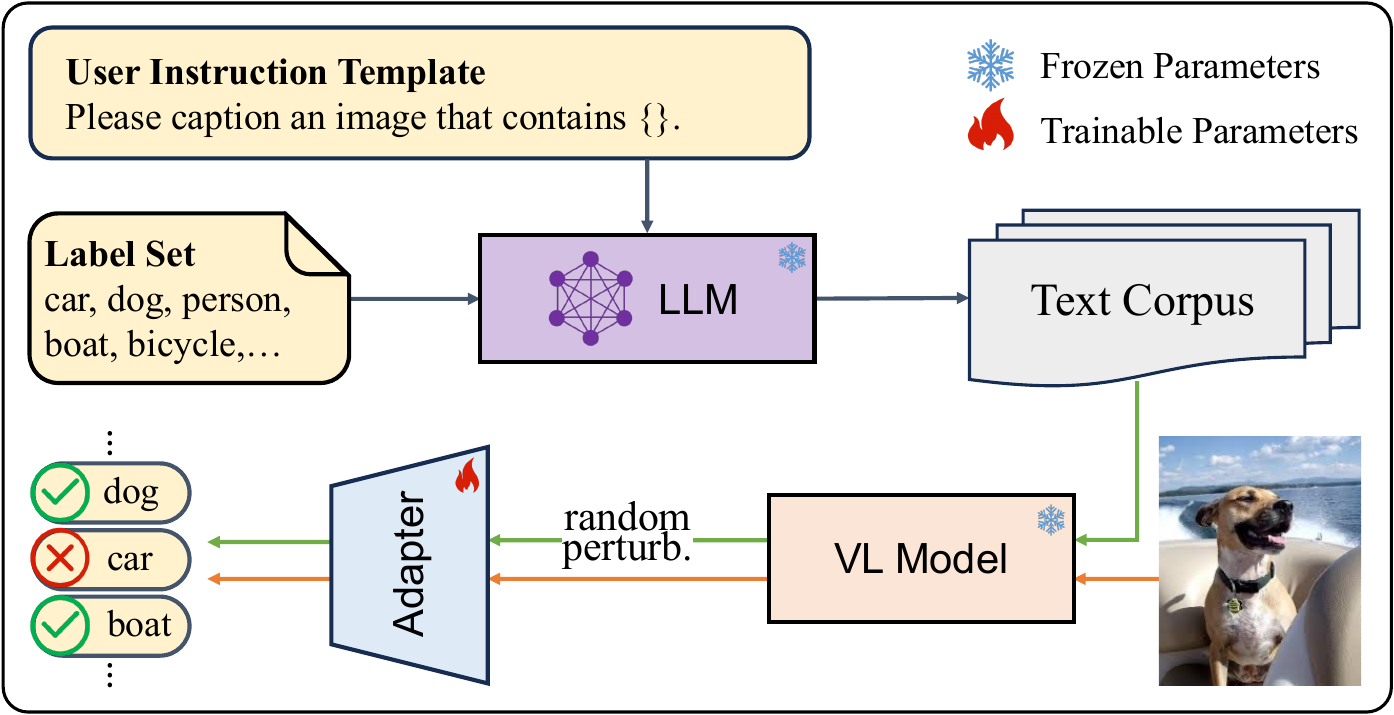}
    \caption{A brief illustration of the TaI-Adapter framework. The green and red lines indicate the training process in text modality and the inference process in visual modality, respectively.}
    \label{fig:pipeline}
\end{figure}

Once the multi-label instruction-following texts are collected, we further propose an advanced framework that takes \underline{T}ext \underline{a}s \underline{I}mage to learn a transferable \underline{Adapter} (TaI-Adapter) for multi-label zero-shot learning (ZSL), as illustrated in Fig. \ref{fig:pipeline}. Specifically, during the training process, texts are first fed into the VL text encoder and the encoded text embeddings are subsequently used to train an adapter network for label recognition. As for the inference process, images are input into the VL image encoder followed by the well-trained adapter network to yield label predictions. To enhance the cross-modal transfer ability of the adapter, random perturbation is introduced to search for potential image embeddings by injecting noise into text embeddings, allowing the adapter trained in text modality to perform better in visual modality. In addition, two variants of random perturbation are further designed to fully utilize available images to improve the performance of the adapter in multi-label few-shot learning (FSL) and partial-label learning (PLL) tasks. Overall, our main contributions are as follows:

\begin{itemize}
    \item We propose a novel framework that takes text as image to learns a transferable adapter in the VL embedding space. By grafting with LLM-driven text generation, this framework provides a fully automated solution to identify any labels of interest without relying on any manual data.
    \item We introduce a random perturbation mechanism as well as its variants to enhance the cross-modal transfer capability of the adapter network, thereby improving its performance in multi-label image classification.
    \item We conduct extensive experiments on multiple public benchmarks to demonstrate the superiority of our method in multi-label zero-shot learning and few-shot learning as well as partial-label learning tasks.
\end{itemize}


\begin{figure*}[t]
    \centering
    \includegraphics[width=1\linewidth]{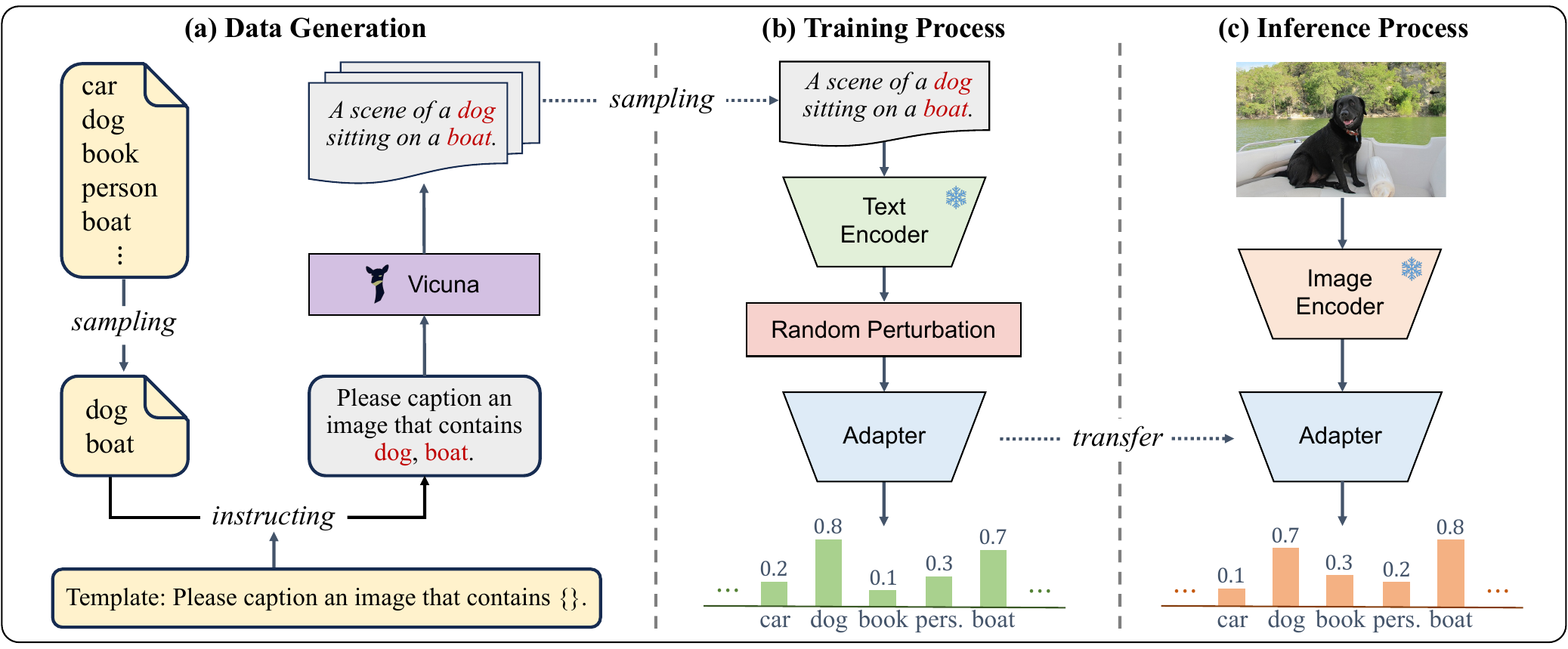}
    \caption{The detailed pipeline of the proposed TaI-Adapter framework. It contains a LLM-driven data generation process that produces texts for the lable set of interest, a training process that employs the generated texts to learn an adapter network, and an inference process that transfers the learned adapter from text modality to visual modality for label recognition.}
    \label{fig:framework}
\end{figure*}

\section{Related Work}

\subsection{Multi-Label Image Classification}
As a fundamental task in computer vision, multi-label image classification \cite{li2014multi, wei2015hcp, wang2016cnn} aims to identify multiple objects or concepts for an image. To this end, many research efforts have been made to model complicated dependencies between labels. Sequential model based methods \cite{wang2016cnn, wang2017multi, chen2018recurrent, chen2018order, yazici2020orderless} resort to recurrent neural networks like RNN and LSTM to implicitly capture spatial relevance or semantic dependency among labels, while other methods \cite{chen2019multi, chen2019learning, ye2020attention, you2020cross, zhu2023scene} employ GNN or GCN to explicitly model label co-occurrence relationships by graph propagation. 

As vision Transformer \cite{dosovitskiy2020image} emerges rapidly, recent work attempts to simultaneously explore spatial relationships and label dependencies as well as their cross-modal alignments in a self-attention or cross-attention manner \cite{liu2021query2label, zhu2022two}. In addition, in view of the characteristics of multi-label image classification task, some customized designs, such as residual attention \cite{zhu2021residual}, asymmetric learning \cite{ridnik2021asymmetric} and group-decoding scheme \cite{ridnik2023ml}, are proposed to further improve the performance. However, these methods heavily rely on a large number of labeled images for model optimization, and could suffer significant performance degradation when encountering data-limited or label-limited scenarios.

\subsection{Multi-Label Zero (Few) -Shot Learning}
Multi-label zero (few) -shot learning is a cross task of multi-label classification and zero (few) -shot learning in computer vision, encountering challenges of both fields. Most early works \cite{huynh2020shared, ben2021semantic, narayan2021discriminative} deploy models trained on images with seen labels to novel labels, while the literatures \cite{alfassy2019laso} and \cite{simon2022meta} propose feature manipulation and meta-learning framework for better exploit the small number of images with novel labels, respectively. Albeit effective, they still require images annotated by seen labels for training. 

Recently, it has become prevalent to transfer the pre-trained knowledge of CLIP \cite{radford2021learning} to downstream vision tasks for data-limited learning. Prompt learning \cite{zhou2022learning, zhou2022conditional, sun2022dualcoop, guo2023texts} has been a dominant approach for zero (few) -shot multi-label classification in an open-vocabulary way. Although considerable progress towards open-world recognition has been made, they suffer from heavy computation costs and memory consumption, and even become computational prohibitive when label size grows larger.

\subsection{Multi-Label Partial-Label Learning}

Multi-label partial-label learning is a challenging task with only partial labels for each image being known for training. To enable such kind of label-limited learning, partial-BCE \cite{durand2019learning} generalizes the standard binary cross-entropy loss to adapt to the proportion of known labels per image. SST \cite{chen2022structured} and SARB \cite{pu2022semantic} propose to complement unknown labels by exploring the semantic correlations of images and their interactions with class proxies, respectively. Some recent attempts \cite{sun2022dualcoop, guo2023texts} also have been made to explore the pre-trained knowledge in vision-language models by prompt learning for partial-label learning. Although considerable progress has been made, the exceedingly low model efficiency still persists as a concern. 

\section{Proposed Method}

\begin{figure*}
  \centering
  \begin{subfigure}{0.33\linewidth}
    \includegraphics[width=1.0\linewidth]{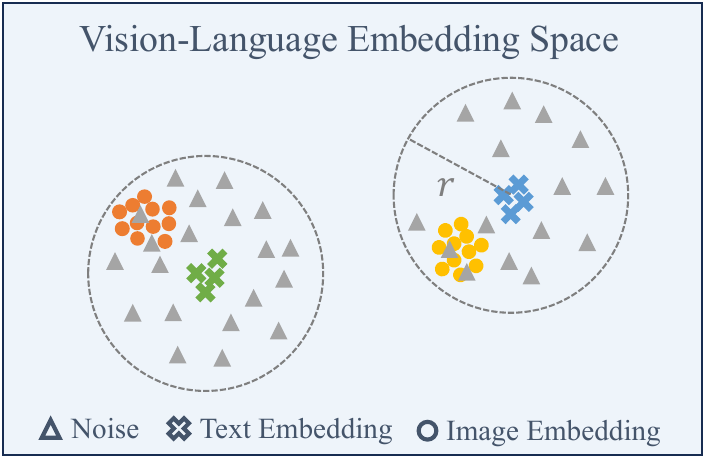}
    \caption{Random perturbation for ZSL.}
    \label{fig:random1}
  \end{subfigure}
  \hfill
  \begin{subfigure}{0.33\linewidth}
    \includegraphics[width=1.0\linewidth]{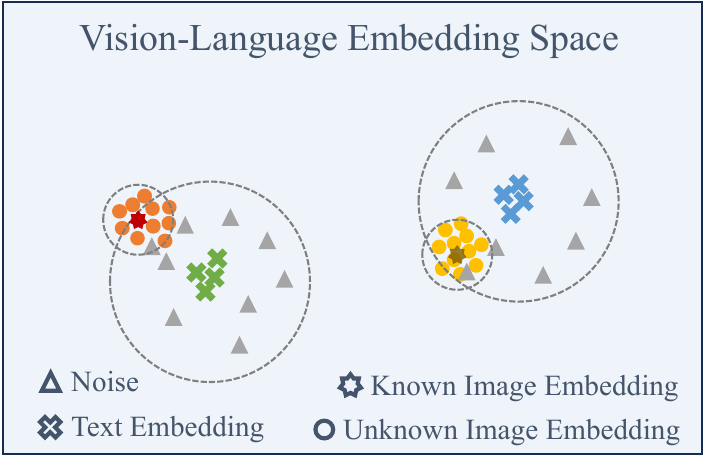}
    \caption{Mixed random perturbation for FSL.}
    \label{fig:random2}
  \end{subfigure}
  \hfill
  \begin{subfigure}{0.33\linewidth}
    \includegraphics[width=1.0\linewidth]{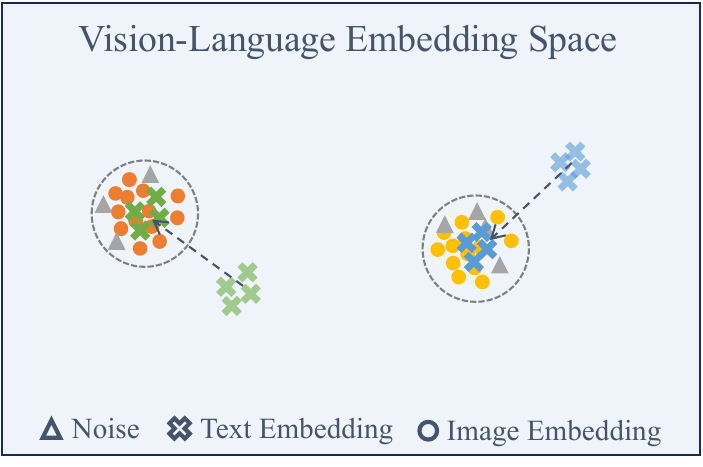}
    \caption{Shifted random perturbation for PLL.}
    \label{fig:random3}
  \end{subfigure}
  \caption{Illustration of random perturbation as well as its mixed and shifted counterparts for multi-label zero-shot learning and few-shot learning as well as partial-label learning tasks, respectively.}
  \label{fig:random}
\end{figure*}

\subsection{Overview}
The detailed pipeline of the whole TaI-Adapter framework is illustrated in Fig. \ref{fig:framework}. As shown, for a given candidate label set, the data generation module leverages the LLM-based chatbot Vicuna \cite{chiang2023vicuna} to produce a large number of multi-label instruction-following texts. During the training process, these texts are encoded by the CLIP text encoder to train an adapter network for label recognition. Then, in inference process, the input image is first fed into the CLIP image encoder, and the output visual embedding is subsequently input into the well-trained adapter network for visual label prediction. Specifically, random perturbation is introduced to enhance the cross-modal transfer capacity of the adapter for improved performance in multi-label classification.


\subsection{LLM-Driven Data Generation}




In this section, we introduce how to employ large language models to generate multi-label instruction-following texts for the label set of interest. The detailed pipeline of LLM-driven data generation is illustrated in the left part in Fig. \ref{fig:framework}. Specifically, Vicuna \cite{vicuna2023}, an open-source chatbot trained by fine-tuning the large language model LLaMA \cite{touvron2023llama} on user-shared conversations, is selected for multi-label instruction-following text generation. For a given candidate label set of interest, we first randomly sample several labels and populate them into a well-designed instruction template, \textit{e.g.}, ``\textit{Please briefly caption an image that contains \{\}.}", yielding a label-specific instruction. Then, the instruction is employed to drive the Vicuna to generate related texts. After that, some optional pre-set instructions can be used to further refine the generated texts to meet some requirements (length, diversity, etc.) in a conversational way. Several conversations with Vicuna in practice have been exhibited in appendix, please refer to it for more details. In this way, without any manual involvement, we are able to effortlessly collect a substantial quantity of high-quality texts that cover a variety of label combinations for the subsequent training of the adapter network.


\subsection{Text as Image for Multi-Label Classification}



For convenience, we first denote the candidate label set as $\mathcal{L}=\{l_0,l_1,\cdots,l_{N-1}\}$ with $N$ being the number of labels. The collected multi-label instruction-following text data are denoted as $\mathcal{D}=\{(T_i,\mathbf{y}_i)\}_{i=0}^{M-1}$, where $M$ is the number of texts; $\mathbf{y}_i=\{y_{i0},y_{i1},\cdots,y_{i,N-1}\}$ denotes the ground truth of the text $T_i$ and $y_{ij}$ for $j\in\{0,1,\cdots,N-1\}$ is $1$ if the text $T_i$ is generated from the label $l_j$ and $0$ otherwise. Then, the text $T_i$ is input into the CLIP text encoder to produce its text embedding, formulated as follows:
\begin{equation}
    \mathbf{t}_i = \Phi_{\mathrm{T}}(T_i),
\end{equation}
where $\Phi_{\mathrm{T}}(\cdot)$ denotes the CLIP text encoder and $\mathbf{t}_i\in\mathbb{R}^d$ with $d$ being the dimension of the text embedding. Then, an adapter network is designed to compute the confidence scores of the text $T_i$ over all labels, formulated as:
\begin{equation}
    \mathbf{p}_i = \sigma(\Phi_\mathrm{A}(\mathbf{t}_i)),
\end{equation}
where $\sigma(\cdot)$ is the sigmoid function and $\Phi_\mathrm{A}(\cdot)$ denotes the adapter network; $\mathbf{p}_i=\{p_{i0},p_{i1},\cdots,p_{i,N-1}\}$ with $p_{ij}$ for $j\in\{0,1,\cdots,N-1\}$ being the probability that the text $T_i$ is generated from the label $l_j$. Finally, the binary cross entropy loss is calculated for optimization:
\begin{equation}
    L = -\frac{1}{N}\sum_{j=0}^{N-1}y_{ij}\log p_{ij}+(1-y_{ij})\log(1-p_{ij}).
\end{equation}

Once the adapter network is well trained, it is able to be directly transferred from the text modality into the visual modality in the CLIP embedding space. Specifically, providing an input image $I$, we first input it into the CLIP image encoder to produce its image embedding:
\begin{equation}
    \mathbf{v} = \Phi_\mathrm{V}(I),
\end{equation}
where $\Phi_\mathrm{V}(\cdot)$ denotes the CLIP image encoder and $\mathbf{v}\in\mathbb{R}^d$. Then, similar to text modality, the encoded image embedding is fed into the adapter network $\Phi_\mathrm{A}(\cdot)$ followed by a sigmoid function for label prediction.

As a result, a fully automated framework, namely TaI-Adapter, is developed for multi-label image classification. For any set of labels of interest, it employs LLM to generate multi-label text data, and then trains a transferable adapter network in text modality to achieve visual label recognition.

\subsection{Random Perturbation for ZSL}

Although the embeddings of texts and images with similar semantics are able to be aligned in the CLIP embedding space, they are unlikely to overlap exactly due to the modality gap. This may result in the adapter network learned from text modality not being able to transfer well to visual modality, limiting its performance in multi-label ZSL task. Nonetheless, there is no doubt that in the CLIP embedding space, the embeddings of images and texts are closer to each other if they have the same labels, otherwise they are farther apart, as depicted in Fig. \ref{fig:random1}.

Based on the above analysis, we introduce a simple yet effective approach, termed random perturbation, to enhance the cross-modal transfer ability of the adapter network by perturbing text embeddings with noise. Concretely, with the encoded text embedding $\mathbf{t}_i$, we randomly sample a noise vector $\mathbf{\epsilon}\in\mathbb{R}^d$ from the hypersphere centered at the origin with a radius of $r$ and inject it into the text embedding $\mathbf{t}_i$:
\begin{equation}
    \mathbf{t}'_i = \mathbf{t}_i + \mathbf{\epsilon},
\end{equation}
After that, the noise-injected text embedding $\mathbf{t}'_i$ serves as an alternative for training the adapter network. In this way, the adapter network is able to perceive underlying visual embeddings around text embeddings during the training phase, thereby better transferring from text modality to visual modality for multi-label ZSL. On the other hand, labels commonly appear explicitly in texts, and the exposure of supervision signals makes the adapter network prone to overfitting to text modality. Random perturbation can effectively increase the diversity of text embeddings, thereby preventing the adapter from overfitting.

\subsection{Mixed Random Perturbation for FSL}

In multi-label FSL task, a small number of images for each label are available during training. To fully leverage these precious labeled images to improve the performance of the adapter network in label recognition, we propose a mixed random perturbation mechanism, which performs perturbation in text modality and visual modality respectively by randomly sampling noise within a hypersphere of preferred radius, as shown in Fig \ref{fig:random2}.


For convenience, providing an image $I'$ that is available during training under the $n$-shot setting, we first input it into the CLIP image encoder to obtain its visual embedding, which is subsequently perturbed by randomly sampling a noise $\epsilon'$ inside a hypersphere of radius $r'$, formulated as:
\begin{equation}
    \mathbf{v}'_k = \Phi_\mathrm{V}(I') + \epsilon'.
\end{equation}
Note that the sampling radius $r'$ in visual modality is much smaller than the radius $r$ in text modality. In this way, more potential images with the same labels as $I'$ are able to be searched. In practice, both the noise-injected text embeddings and $n$-shot image embeddings are used together to train a robust adapter network. 

\subsection{Shifted Random Perturbation for PLL}

In multi-label PLL task, only partial labels for each image are known during training. With this setting, we propose a shifted random perturbation mechanism, which aims to shift text embeddings closer to the clusters of image embeddings that have the same labels as them, allowing the subsequent random perturbation on text embeddings to cover potential visual embeddings easily, as shown in Fig. \ref{fig:random3}.

Without loss of generality, we denote the whole image dataset as $\{(I_k,\widetilde{\mathbf{y}}_k)\}_{k=0}^{K-1}$, where $K$ is the number of images; $\widetilde{\mathbf{y}}_k$ indicates the multi-hot annotation of the image $I_k$ and $\widetilde{\mathbf{y}}_k=\{\widetilde{y}_{k0},\widetilde{y}_{k1},\cdots,\widetilde{y}_{k,N-1}\}$. Following the general setting of PLL task, $\widetilde{y}_{kj}$ for $j\in\{0,1,\cdots,N-1\}$ is assigned to 1 if the label $l_j$ exists in the image $I_k$, to 0 if it does not exist, and to -1 if it is unknown. Then, all images are first fed into the CLIP image encoder to produce their visual embeddings, which are subsequently decoupled to obtain label-specific visual embeddings, formulated as:
\begin{equation}
    \mathbf{c}^\mathrm{V}_j =\frac{\sum_{k=0}^{K-1}\mathds{1}(\widetilde{y}_{kj}>0)\Phi_\mathrm{V}(I_k)}{\sum_{k=0}^{K-1}\mathds{1}(\widetilde{y}_{kj}>0)},
\end{equation}
where $\mathds{1}(\cdot)$ is an indicator function; $\mathbf{c}^\mathrm{V}_j\in\mathbb{R}^d$ is the visual embedding of label $l_j$. Then, for any label combination, we computer its centroid vector in text modality by averaging all corresponding text embeddings. Taking $\mathbf{y}_i$ as an example, its text embedding set denotes as $\mathcal{T}_i$, and the centroid vector is calculated as follows:
\begin{equation}
    \mathbf{c}^\mathrm{T}_i = \frac{1}{|\mathcal{T}_i|}\sum_{t=0}^{|\mathcal{T}_i|-1}\mathbf{t}_t.
\end{equation}
Then, the offset vector of the label combination $\mathbf{y}_i$ between text modality and visual modality can be obtained:
\begin{equation}
    \mathbf{o}_i = \frac{\sum_{j=0}^{N-1}y_{ij}\mathbf{c}^\mathrm{V}_j}{\sum_{j=0}^{N-1}y_{ij}} - \mathbf{c}^\mathrm{T}_i.
\end{equation}
Once computed, we are able to shift the text embedding $\mathbf{t}_i$ closer to the corresponding image clusters by:
\begin{equation}
    \widetilde{\mathbf{t}}_i = \mathbf{t}_i + \mathbf{o}_i.
\end{equation}
After that, random perturbation is performed to inject noise into the shifted text embedding $\widetilde{\mathbf{t}}_i$ for training the adapter network. With such shift mechanism, text embeddings are able to be moved towards the clusters of the corresponding image embeddings, shortening the gap between them, even overlapping with each other. Consequentially, the potential visual embeddings around text embeddings are more easily to be sampled by random perturbation, enabling the trained adapter network to perform better in visual modality.

\renewcommand{\tabcolsep}{0pt}
\newcolumntype{Y}{p{1.18cm}<{\centering}}

\begin{table*}[t]
 \small
\begin{tabularx}{\textwidth}{X<{\centering}|p{3.0cm}<{\centering}|YYYYYYYYY|Y}
\toprule
Dataset & Method & 10\% & 20\% & 30\% & 40\% & 50\% & 60\% & 70\% & 80\% & 90\% & Avg. \\
\hline
\hline
\multirow{4}{*}{VOC 2007} & SARB \cite{pu2022semantic} & 83.5 & 88.6 & 90.7 & 91.4 & 91.9 & 92.2 & 92.6 & 92.8 & 92.9 & 90.7 \\
 & DualCoOp \cite{sun2022dualcoop} & 91.4 & 93.8 & 93.8 & 94.3 & 94.6 & 94.7 & 94.8 & 94.9 & 94.9 & 94.1 \\
 & TaI-DPT* \cite{guo2023texts} & 93.3 & 94.6 & 94.8 & 94.9 & 95.1 & 95.0 & 95.1 & 95.3 & 95.5 & 94.8 \\
 & \textbf{TaI-Adapter}* & \textbf{93.8} & \textbf{94.7} & \textbf{95.1} & \textbf{95.2} & \textbf{95.3} & \textbf{95.3} & \textbf{95.4} & \textbf{95.6} & \textbf{95.7} & \textbf{95.1} \\
\hline
\hline
\multirow{4}{*}{MS-COCO} & SARB \cite{pu2022semantic} & 71.2 & 75.0 & 77.1 & 78.3 & 78.9 & 79.6 &79.8 & 80.5 & 80.5 & 77.9 \\
 & DualCoOp \cite{sun2022dualcoop} & 81.0 & 82.3 & 82.9 & 83.4 & 83.5 & 83.9 & 84.0 & 84.1 & 84.3 & 83.3 \\
 & TaI-DPT* \cite{guo2023texts} & 81.5 & 82.6 & 83.3 & 83.7 & 83.9 & 84.0 & 84.2 & 84.4 & 84.5 & 83.6 \\
& \textbf{TaI-Adapter}* & \textbf{82.1} & \textbf{82.9} & \textbf{83.5} & \textbf{84.0} & \textbf{84.4} & \textbf{84.7} & \textbf{84.9} & \textbf{85.1} & \textbf{85.1} & \textbf{84.1} \\
\hline
\hline
\multirow{3}{*}{NUS-WIDE} & DualCoOp \cite{sun2022dualcoop} & 54.0 & 56.2 & 56.9 & 57.4 & 57.9 & 57.9 & 57.6 & 58.2 & 58.8 & 57.2 \\
 & TaI-DPT* \cite{guo2023texts} & 56.4 & 57.9 & 57.8 & 58.1 & 58.5 & 58.8 & 58.6 & 59.1 & 59.4 & 58.3 \\
 & \textbf{TaI-Adapter}* & \textbf{59.5} & \textbf{61.9} & \textbf{62.8} & \textbf{63.2} & \textbf{63.2} & \textbf{63.1} & \textbf{63.5} & \textbf{64.2} & \textbf{65.0} & \textbf{62.9} \\
\bottomrule
\end{tabularx}
\caption{Performance comparison of our TaI-Adapter and existing methods in multi-label partial-label learning task (mAP in \%). The symbol * indicates that the method reports results by integrating with DualCoOp.}
\label{tab:partial-label}
\end{table*}

\newcolumntype{Y}{p{2.0cm}<{\centering}}
\begin{table}[t]
\small
\centering
\begin{tabularx}{\linewidth}{X<{\centering}|Y|Y|Y}
\toprule
Method & VOC 2007 & MS-COCO & NUS-WIDE \\
\hline
\hline
CLIP \cite{radford2021learning} & 77.3 & 49.7 & 37.4 \\
 \hline
TaI-DPT \cite{guo2023texts} & 88.3 & 65.1 & 46.5 \\
 \hline
\textbf{TaI-Adapter} & \textbf{89.0} & \textbf{67.7} & \textbf{53.3} \\
\bottomrule
\end{tabularx}
\caption{Performance comparison of our TaI-Adapter and existing methods in multi-label zero-shot learning task (mAP in \%).}
\label{tab:zero-shot}
\end{table}

\section{Experiments}

\subsection{Experiment Setup}

\textbf{Network Architecture.} In this work, we choose the CLIP ResNet-50 \cite{he2016deep} as the image encoder and the build-in Transformer as the text encoder following previous work \cite{guo2023texts}. The Vicuna-33b-1.3v is selected for multi-label instruction-following data generation. The adapter network is a three-layer feed-forward neural network with each layer consisting of a fully-connected layer followed by a ReLU activation and a dropout layer \cite{srivastava2014dropout}. During training, all parameters of the image encoder and text encoder are frozen and only the adapter network is trainable.

\noindent\textbf{Datasets.} We conduct experiments on three public benchmarks, including Pascal VOC 2007 \cite{everingham2010pascal}, Microsoft COCO (MS-COCO) \cite{lin2014microsoft} and NUS-WIDE \cite{chua2009nus}. Specifically, Pascal VOC 2007 \cite{everingham2010pascal} has 20 label categories in total and 9,963 images, in which 5,011 images form \textit{train-val} set and remaining 4,952 images are taken as \textit{test} set for evaluation. MS-COCO \cite{lin2014microsoft} contains 82,081 images for the training set and 40,137 images for the validation set, and covers 80 label categories. NUS-WIDE \cite{chua2009nus} is a web dataset with 161,789 images for training and remaining 107,859 images for testing, covering 81 human verified labels.

\noindent\textbf{Implementation Details.}  The whole framework is trained for 60 epochs using AdamW \cite{loshchilov2017decoupled} optimizer with a batch size of 64 and 1-cycle policy \cite{smith2018disciplined} with a maximum learning rate of 0.0001. For a fair comparison with existing methods, the input images are resized into 224$\times$224 in the multi-label ZSL and FSL tasks for evaluation, and 448$\times$448 in the multi-label PLL task for both training and evaluation. The radius is set as 25, 1 and 10 for random perturbation as well as its mixed and shifted counterparts respectively.


\subsection{Comparison with State-of-the-arts}

\textbf{Results on Multi-Label ZSL Task.} To evaluate the effectiveness of our proposed TaI-Adapter framework, we compare its performance with existing methods, including CLIP \cite{radford2021learning} and TaI-DPT \cite{guo2023texts}. Experimental results on three public multi-label datasets are reported in Table \ref{tab:zero-shot}. As shown, our TaI-Adapter consistently achieves best performance, leading TaI-DPT by a considerable margin of 0.7\%, 1.3\% and 6.8\% in mAP on Pascal VOC 2007, MS-COCO and NUS-WIDE, respectively. Notably, our TaI-Adapter shows more powerful capability in the multi-label ZSL task. It should be noted that our TaI-Adapter is a fully automated framework for novel label recognition without relying on any manual data, whether images or texts, making it more valuable for real-world applications. More importantly, our TaI-Adapter has faster training speeds, lower GPU memory consumption compared to the prompt learning based methods, as shown in Fig. \ref{fig:memory}. We refer readers to Sec. \ref{sec:efficiency} for more details.

\noindent\textbf{Results on Multi-Label FSL Task.} We further compare our TaI-Adapter with current methods in multi-label FSL task. Following previous work \cite{alfassy2019laso, guo2023texts}, we randomly sample 1, 2, 4, 8 and 16 -shot images for each label as known samples for model training. The upper and the lower parts of Table \ref{tab:few-shot} present experimental results on VOC 2007 and MS-COCO, respectively. Note that we follow the same setting of TaI-DPT and integrate the predicted scores of TaI-Adapter with CoOp. Obviously, our method achieves consistent performance advantage in various few-shot settings on both datasets, demonstrating its superiority in multi-label FSL task. The performance of TaI-Adapter without integrating with CoOp is discussed in Sec. \ref{sec:perturbation}.

\newcolumntype{Y}{p{1.2cm}<{\centering}}
\begin{table}[t]
\small
\centering
\begin{tabularx}{\linewidth}{X<{\centering}|Y|Y|Y|Y|Y}
\toprule
Method & 1-shot & 2-shot & 4-shot & 8-shot & 16-shot \\
\hline
\hline
CoOp \cite{zhou2022learning} & 79.3 & 83.2 & 83.8 & 84.5 & 85.7 \\
 \hline
TaI-DPT* \cite{guo2023texts} & 89.7 & 91.1 & 91.2 & 92.4 & 93.1 \\
\hline
\textbf{TaI-Adapter}* & \textbf{90.0} & \textbf{91.5} & \textbf{92.2} & \textbf{93.5} & \textbf{93.9} \\
\hline
\hline
CoOp \cite{zhou2022learning} & 52.6 & 57.3 & 58.1 & 59.2 & 59.8 \\
 \hline
TaI-DPT* \cite{guo2023texts} & 70.0 & 70.1 & 70.8 & 71.2 & 71.8 \\
\hline
\textbf{TaI-Adapter}* & \textbf{70.5} & \textbf{70.7} & \textbf{71.4} & \textbf{72.0} & \textbf{72.7} \\
\bottomrule
\end{tabularx}
\caption{Performance comparison of our TaI-Adapter and existing methods in multi-label few-shot learning task (mAP in \%).The upper part and the lower part report results on VOC 2007 and MS-COCO, respectively. The symbol * indicates that the method reports results by integrating with CoOp.}
\label{tab:few-shot}
\end{table}


\begin{figure*}
  \centering
  \begin{subfigure}{0.33\linewidth}
    \includegraphics[width=1.0\linewidth]{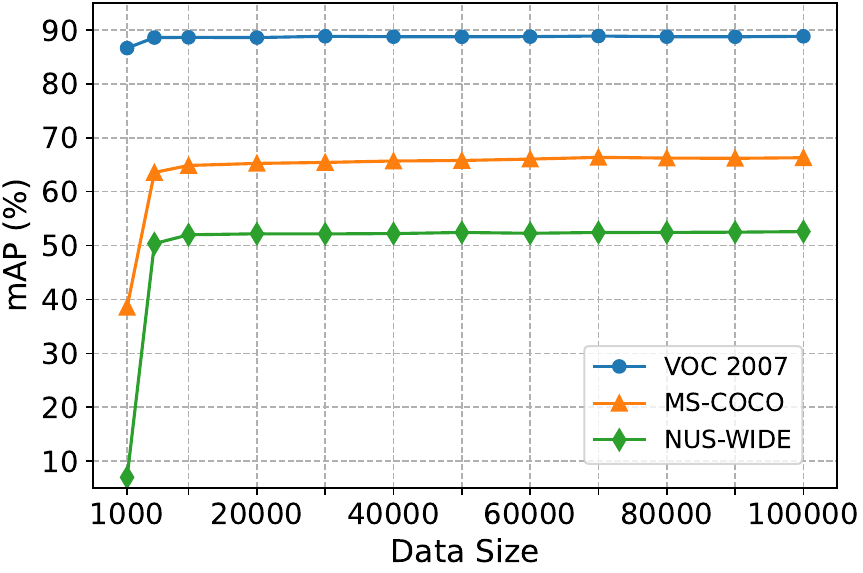}
    \caption{Effect of data size.}
    \label{fig:size}
  \end{subfigure}
  \hfill
  \begin{subfigure}{0.33\linewidth}
    \includegraphics[width=1.0\linewidth]{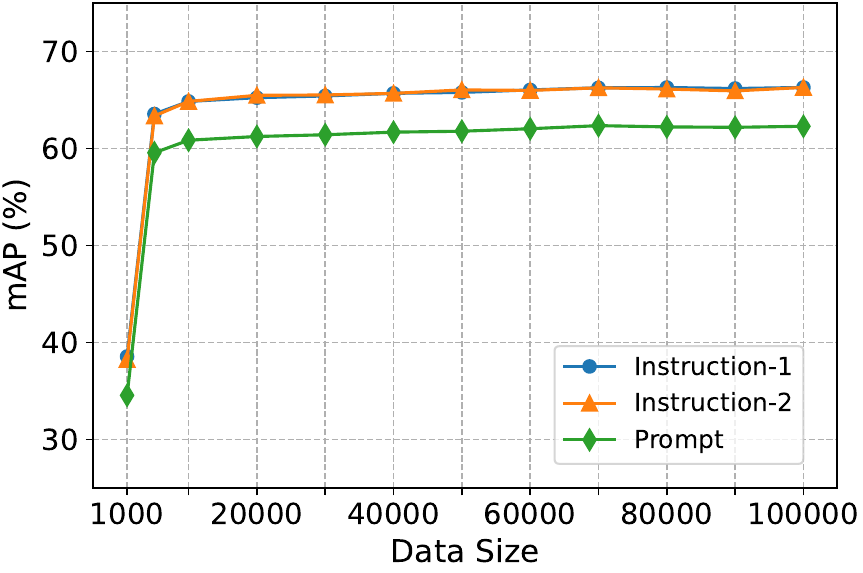}
    \caption{Effect of generation method.}
    \label{fig:template}
  \end{subfigure}
  \hfill
  \begin{subfigure}{0.33\linewidth}
    \includegraphics[width=1.0\linewidth]{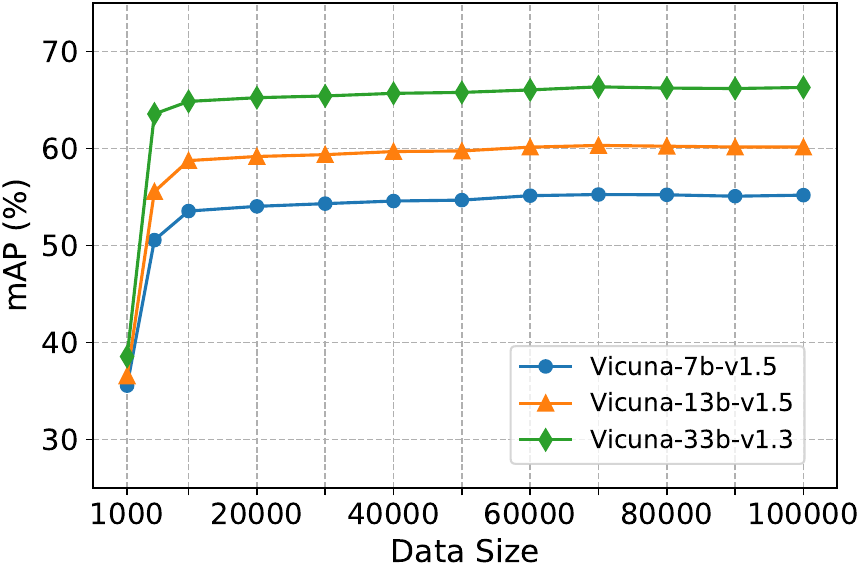}
    \caption{Effect of LLM architecture.}
    \label{fig:arch}
  \end{subfigure}
  \caption{Effect of the LLM-drive instruction-following data generation on the performance of the proposed TaI-Adapter framework.}
  \label{fig:llm}
\end{figure*}

\noindent\textbf{Results on Multi-Label PLL Task.} We also evaluate the proposed TaI-Adapter framework in multi-label PLL task. Table \ref{tab:partial-label} compares its performance with existing methods on VOC 2007, MS-COCO and NUS-WIDE datasets. As shown, under the diverse rates of known labels, our method consistently accomplishes better performance than TaI-DPT with the same setting (integrating results with DualCoOp \cite{sun2022dualcoop}) on all datasets, showing the superiority of our method in multi-label PLL task. We leave the model performance without DualCoOp to be discussed in Sec. \ref{sec:perturbation}.

\subsection{Ablation Study}

\subsubsection{Effect of LLM-Drive Data Generation}

Here we explore the effect of the LLM-drive data generation on the performance of TaI-Adapter in multi-label ZSL.

\noindent\textbf{Effect of Data Size.} We first exploit how data size affects the performance of the adapter network. Experimental results on VOC 2007, MS-COCO and NUS-WIDE are shown in Fig. \ref{fig:size}. Our TaI-Adapter accomplishes impressive performance in multi-label ZSL task while only requiring 10,000 texts for training. In particular, on the VOC 2007 dataset with only 20 labels, an adapter network with decent performance can be trained with only 1,000 texts. Notably, our TaI-Adapter is a data-efficient learning framework that enables rapid implementation of recognition tasks for a small set of labels in practical applications.

\noindent\textbf{Effect of Generation Method.} We further investigate the impact of the data generation method on model performance. To this end, two instruction templates (``Instruction-1: \textit{please briefly caption an image that contains \{\}}." and ``Instruction-2: \textit{please organize the words \{\} into a sentence to describe an image.}") are designed to drive LLM for instruction-following text generation. Also, a set of prompts (\textit{e.g.}, \textit{there are \{\} in the photo.}) is predefined for prompt-based text generation. Fig. \ref{fig:template} shows experimental results on MS-COCO. The mAP curves of the two instruction templates almost completely overlap, suggesting that our model is not sensitive to the diversity of instructions. Besides, despite a certain level of performance degradation, the model trained on prompt-based texts still achieved decent performance. Compared to the LLM-based instruction-following text generation that takes several hours to complete, prompt-based text generation can be done in just a few seconds. This provides a fast solution for visual label recognition.

\noindent\textbf{Effect of LLM Architecture.} We also explore the impact of LLM architecture on model's performance. Hence, Vicuna-33b-v1.3, Vicuna-13b-v1.5 and Vicuna-7b-v1.5 are selected for comparison. Fig. \ref{fig:arch} reports experimental results on MS-COCO. We can see that language models with larger number of parameters exhibit superior performance curves. This is reasonable as larger language models possess stronger language modeling abilities, leading to higher-quality text generation. Consequently, the trained adapter exhibits better performance in visual label recognition.


\begin{figure*}
  \centering
  \begin{subfigure}{0.33\linewidth}
    \includegraphics[width=1.0\linewidth]{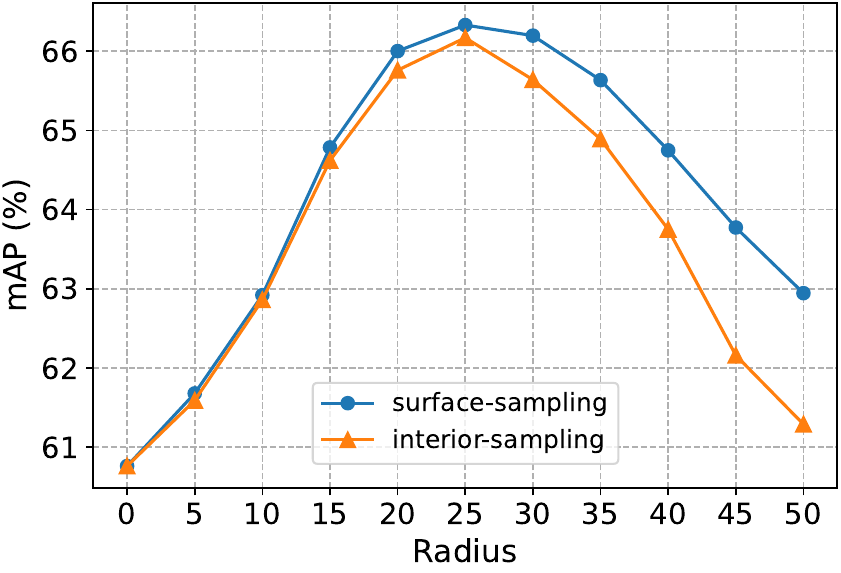}
    \caption{Effect of random perturbation.}
    \label{fig:sampling}
  \end{subfigure}
  \hfill
  \begin{subfigure}{0.33\linewidth}
    \includegraphics[width=1.0\linewidth]{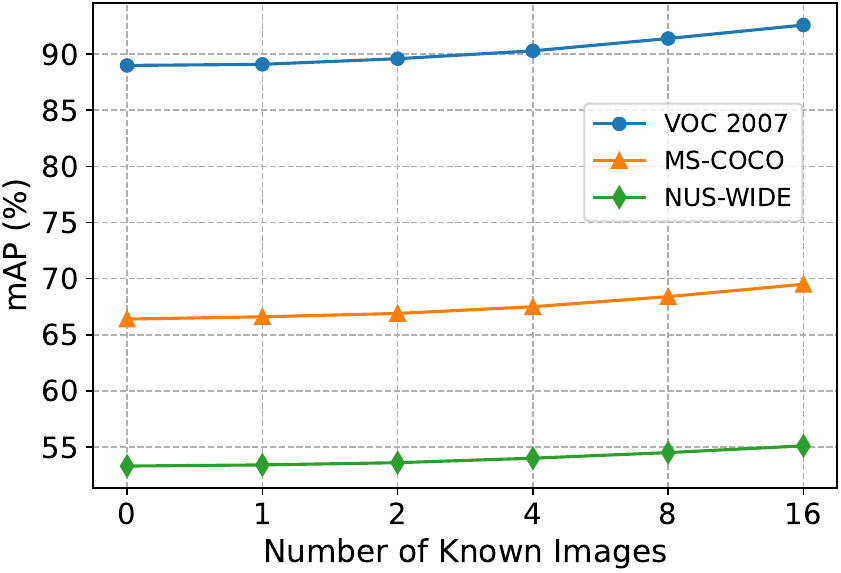}
    \caption{Effect of mixed random perturbation.}
    \label{fig:few-shot}
  \end{subfigure}
  \hfill
  \begin{subfigure}{0.33\linewidth}
    \includegraphics[width=1.0\linewidth]{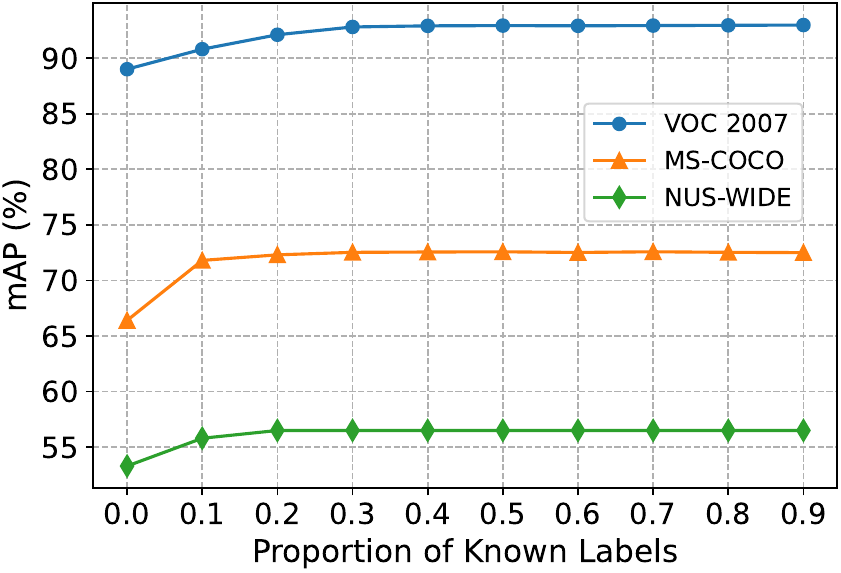}
    \caption{Effect of shifted random perturbation.}
    \label{fig:partial-label}
  \end{subfigure}
  \caption{Effect of the random perturbation mechanism as well as its variants on the performance of the proposed TaI-Adapter framework.}
  \label{fig:perturbation}
\end{figure*}

\begin{figure}
    \centering
    \includegraphics[width=0.8\linewidth]{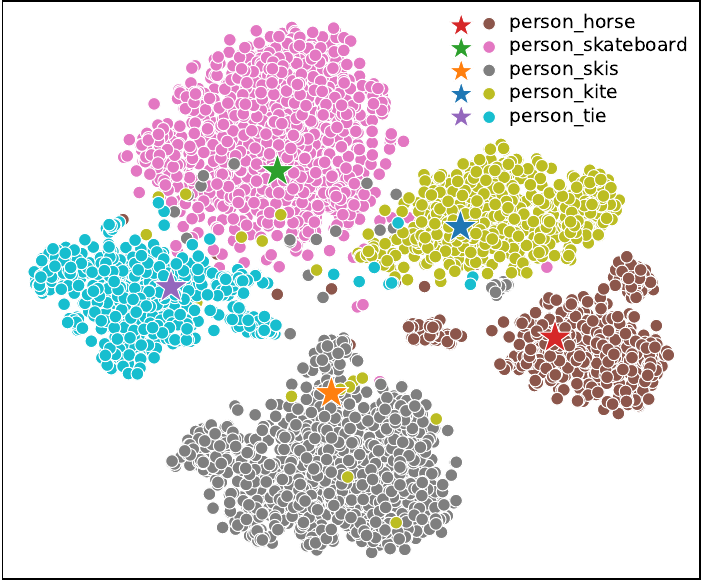}
    \caption{Visualization on the distribution of image embeddings and estimated centroids (indicated with stars) in the CLIP embedding space using t-SNE \cite{van2008visualizing} for dimensionality reduction.}
    \label{fig:tnse}
\end{figure}

\subsubsection{Effect of Random Perturbation \label{sec:perturbation}}

Here we exploit the effect of random perturbation as well as its variants on the performance of TaI-Adapter.

\noindent\textbf{Effect of Random Perturbation.} To figure out how the random perturbation affects the performance of the TaI-Adapter, we explore sampling schemes from the surface and interior of hyperspheres of varying radii. Fig. \ref{fig:sampling} presents experimental results on MS-COCO. As shown, the performance curves climb rapidly as the radius increases for both sampling schemes, and reach their peak when the radius is 25. At this point, the mAP surpasses that at the radius of 0, where no perturbation is performed, by more than 5\%, proving the effectiveness of random perturbation. Besides, surface sampling scheme outperforms interior sampling overall. This could be due to the latter's excessive sampling of noise near the center of the hypersphere, which is failed to search for potential image embeddings.

\noindent\textbf{Effect of Mixed Random Perturbation.} We further explore the effectiveness of the mixed random perturbation in multi-label FSL task. As reported in Fig. \ref{fig:few-shot}, note that the mixed random perturbation degenerates into plain random perturbation when the number of known images is 0. Notably, the performance of our model in various few-shot learning settings consistently outperforms its performance in zero-shot learning; On the other hand, as the number of known images increases, the model's performance exhibits a stable improvement. Therefore, the effectiveness of mixed random perturbation in data-limited learning is proved.

\noindent\textbf{Effect of Shifted Random Perturbation.} We also investigate the effectiveness of the shifted random perturbation in multi-label PLL task. Fig. \ref{fig:partial-label} exhibits experimental results on three datasets. As shown, it can be observed that all mAP curves initially show an upward trend, followed by a gradual stabilization as the proportion of known labels increases further. Overall, the model suffers considerable performance degradation at a proportion of known labels of 0, where the shifted random perturbation is removed, thus demonstrating its effectiveness in label-limited learning. We further confirm this by visualizing the distribution of visual embeddings and the estimated centroids in the CLIP embedding space. As depicted in Fig. \ref{fig:tnse}, our estimated centroids, although slightly deviating from the true centroids, are able to provide an approximate location of image clusters and guide the shifted random perturbation to effectively shift text embeddings closer, allowing the trained adapter to better transfer to the visual modality.


\subsection{Comparison on Model Efficiency \label{sec:efficiency}}

In this section, we compare the efficiency of our model with the state-of-the-art model, \textit{i.e.}, TaI-DPT \cite{guo2023texts}. Results on MS-COCO dataset with 80 labels are reported in Table \ref{tab:efficiency}. As shown, our model has fewer parameters than TaI-DPT, both for training and inference. Especially at inference time, TaI-DPT suffers from significant parameter growth due to the usage of both the image encoder and the text encoder of CLIP. In contrast, the parameter count of our model remains almost unchanged from training. Furthermore, TaI-DPT has over 156$\times$ and 57$\times$ more FLOPs than our model during training and inference, respectively. This is mainly due to the fact that the former requires encoding all prompt texts for each label, resulting in a dramatic increase in computational effort. Clearly, our TaI-Adapter exhibits overwhelming advantages in terms of efficiency.

\newcolumntype{Y}{p{1.5cm}<{\centering}}

\begin{table}[t]
\small
\centering
\begin{tabularx}{\linewidth}{X<{\centering}|Y|Y|Y|Y}
\toprule
\multirow{2}{*}{Method} & \multicolumn{2}{c|}{Training} & \multicolumn{2}{c}{Inference} \\
\cline{2-5}
 & \#param. & FLOPs & \#param. & FLOPs \\
\hline
TaI-DPT \cite{guo2023texts} & 50.4M & 624G & 97.4M & 631G \\
 \hline
TaI-Adapter & 35.9M & 4.0G & 34.2M & 10.9G \\
\bottomrule
\end{tabularx}
\caption{Efficiency Comparison of our TaI-Adapter with TaI-DPT.}
\label{tab:efficiency}
\end{table}

\section{Conclusion}
In this work, we propose a fully automated paradigm for multi-label recognition in the CLIP embedding space. It consists of a LLM-driven data generation process that produces multi-label instruction-following texts for the label set of interest, a training process that takes the produced texts as images to train an adapter network, and an inference process that transfers the learned adapter from text modality to visual modality for label recognition. The random perturbation mechanism and its variants are proposed to enhance the cross-modal transfer ability of the adapter in both data-limited learning and label-limited learning. Extensive experiments demonstrate the superiority of our method in terms of both performance and efficiency.

{
    \small
    \bibliographystyle{ieeenat_fullname}
    \bibliography{main}

\begin{thebibliography}{47}
\providecommand{\natexlab}[1]{#1}
\providecommand{\url}[1]{\texttt{#1}}
\expandafter\ifx\csname urlstyle\endcsname\relax
  \providecommand{\doi}[1]{doi: #1}\else
  \providecommand{\doi}{doi: \begingroup \urlstyle{rm}\Url}\fi

\bibitem[Alfassy et~al.(2019)Alfassy, Karlinsky, Aides, Shtok, Harary, Feris,
  Giryes, and Bronstein]{alfassy2019laso}
Amit Alfassy, Leonid Karlinsky, Amit Aides, Joseph Shtok, Sivan Harary, Rogerio
  Feris, Raja Giryes, and Alex~M Bronstein.
\newblock Laso: Label-set operations networks for multi-label few-shot
  learning.
\newblock In \emph{Proceedings of the IEEE/CVF conference on computer vision
  and pattern recognition}, pages 6548--6557, 2019.

\bibitem[Ben-Cohen et~al.(2021)Ben-Cohen, Zamir, Ben-Baruch, Friedman, and
  Zelnik-Manor]{ben2021semantic}
Avi Ben-Cohen, Nadav Zamir, Emanuel Ben-Baruch, Itamar Friedman, and Lihi
  Zelnik-Manor.
\newblock Semantic diversity learning for zero-shot multi-label classification.
\newblock In \emph{Proceedings of the IEEE/CVF International Conference on
  Computer Vision}, pages 640--650, 2021.

\bibitem[Chen et~al.(2018{\natexlab{a}})Chen, Chen, Yeh, and
  Wang]{chen2018order}
Shang-Fu Chen, Yi-Chen Chen, Chih-Kuan Yeh, and Yu-Chiang~Frank Wang.
\newblock Order-free rnn with visual attention for multi-label classification.
\newblock In \emph{Thirty-Second AAAI Conference on Artificial Intelligence},
  2018{\natexlab{a}}.

\bibitem[Chen et~al.(2018{\natexlab{b}})Chen, Wang, Li, and
  Lin]{chen2018recurrent}
Tianshui Chen, Zhouxia Wang, Guanbin Li, and Liang Lin.
\newblock Recurrent attentional reinforcement learning for multi-label image
  recognition.
\newblock In \emph{Proceedings of the AAAI conference on artificial
  intelligence}, 2018{\natexlab{b}}.

\bibitem[Chen et~al.(2019{\natexlab{a}})Chen, Xu, Hui, Wu, and
  Lin]{chen2019learning}
Tianshui Chen, Muxin Xu, Xiaolu Hui, Hefeng Wu, and Liang Lin.
\newblock Learning semantic-specific graph representation for multi-label image
  recognition.
\newblock In \emph{Proceedings of the IEEE/CVF International Conference on
  Computer Vision}, pages 522--531, 2019{\natexlab{a}}.

\bibitem[Chen et~al.(2022)Chen, Pu, Wu, Xie, and Lin]{chen2022structured}
Tianshui Chen, Tao Pu, Hefeng Wu, Yuan Xie, and Liang Lin.
\newblock Structured semantic transfer for multi-label recognition with partial
  labels.
\newblock In \emph{Proceedings of the AAAI conference on artificial
  intelligence}, pages 339--346, 2022.

\bibitem[Chen et~al.(2019{\natexlab{b}})Chen, Wei, Wang, and
  Guo]{chen2019multi}
Zhao-Min Chen, Xiu-Shen Wei, Peng Wang, and Yanwen Guo.
\newblock Multi-label image recognition with graph convolutional networks.
\newblock In \emph{Proceedings of the IEEE/CVF Conference on Computer Vision
  and Pattern Recognition}, pages 5177--5186, 2019{\natexlab{b}}.

\bibitem[Chiang et~al.(2023{\natexlab{a}})Chiang, Li, Lin, Sheng, Wu, Zhang,
  Zheng, Zhuang, Zhuang, Gonzalez, Stoica, and Xing]{vicuna2023}
Wei-Lin Chiang, Zhuohan Li, Zi Lin, Ying Sheng, Zhanghao Wu, Hao Zhang, Lianmin
  Zheng, Siyuan Zhuang, Yonghao Zhuang, Joseph~E. Gonzalez, Ion Stoica, and
  Eric~P. Xing.
\newblock Vicuna: An open-source chatbot impressing gpt-4 with 90\%* chatgpt
  quality, 2023{\natexlab{a}}.

\bibitem[Chiang et~al.(2023{\natexlab{b}})Chiang, Li, Lin, Sheng, Wu, Zhang,
  Zheng, Zhuang, Zhuang, Gonzalez, et~al.]{chiang2023vicuna}
Wei-Lin Chiang, Zhuohan Li, Zi Lin, Ying Sheng, Zhanghao Wu, Hao Zhang, Lianmin
  Zheng, Siyuan Zhuang, Yonghao Zhuang, Joseph~E Gonzalez, et~al.
\newblock Vicuna: An open-source chatbot impressing gpt-4 with 90\%* chatgpt
  quality.
\newblock \emph{See https://vicuna. lmsys. org (accessed 14 April 2023)},
  2023{\natexlab{b}}.

\bibitem[Chua et~al.(2009)Chua, Tang, Hong, Li, Luo, and Zheng]{chua2009nus}
Tat-Seng Chua, Jinhui Tang, Richang Hong, Haojie Li, Zhiping Luo, and Yantao
  Zheng.
\newblock Nus-wide: a real-world web image database from national university of
  singapore.
\newblock In \emph{Proceedings of the ACM international conference on image and
  video retrieval}, pages 1--9, 2009.

\bibitem[Dosovitskiy et~al.(2020)Dosovitskiy, Beyer, Kolesnikov, Weissenborn,
  Zhai, Unterthiner, Dehghani, Minderer, Heigold, Gelly,
  et~al.]{dosovitskiy2020image}
Alexey Dosovitskiy, Lucas Beyer, Alexander Kolesnikov, Dirk Weissenborn,
  Xiaohua Zhai, Thomas Unterthiner, Mostafa Dehghani, Matthias Minderer, Georg
  Heigold, Sylvain Gelly, et~al.
\newblock An image is worth 16x16 words: Transformers for image recognition at
  scale.
\newblock \emph{arXiv preprint arXiv:2010.11929}, 2020.

\bibitem[Durand et~al.(2019)Durand, Mehrasa, and Mori]{durand2019learning}
Thibaut Durand, Nazanin Mehrasa, and Greg Mori.
\newblock Learning a deep convnet for multi-label classification with partial
  labels.
\newblock In \emph{Proceedings of the IEEE/CVF conference on computer vision
  and pattern recognition}, pages 647--657, 2019.

\bibitem[Everingham et~al.(2010)Everingham, Van~Gool, Williams, Winn, and
  Zisserman]{everingham2010pascal}
Mark Everingham, Luc Van~Gool, Christopher~KI Williams, John Winn, and Andrew
  Zisserman.
\newblock The pascal visual object classes (voc) challenge.
\newblock \emph{International journal of computer vision}, 88\penalty0
  (2):\penalty0 303--338, 2010.

\bibitem[Guo et~al.(2023)Guo, Dong, Ji, Bai, Guo, and Zuo]{guo2023texts}
Zixian Guo, Bowen Dong, Zhilong Ji, Jinfeng Bai, Yiwen Guo, and Wangmeng Zuo.
\newblock Texts as images in prompt tuning for multi-label image recognition.
\newblock In \emph{Proceedings of the IEEE/CVF Conference on Computer Vision
  and Pattern Recognition}, pages 2808--2817, 2023.

\bibitem[He et~al.(2016)He, Zhang, Ren, and Sun]{he2016deep}
Kaiming He, Xiangyu Zhang, Shaoqing Ren, and Jian Sun.
\newblock Deep residual learning for image recognition.
\newblock In \emph{Proceedings of the IEEE conference on computer vision and
  pattern recognition}, pages 770--778, 2016.

\bibitem[Huang et~al.(2023)Huang, Zhang, Ma, Tian, Feng, Zhang, Li, Guo, and
  Zhang]{huang2023tag2text}
Xinyu Huang, Youcai Zhang, Jinyu Ma, Weiwei Tian, Rui Feng, Yuejie Zhang,
  Yaqian Li, Yandong Guo, and Lei Zhang.
\newblock Tag2text: Guiding vision-language model via image tagging.
\newblock \emph{arXiv preprint arXiv:2303.05657}, 2023.

\bibitem[Huynh and Elhamifar(2020)]{huynh2020shared}
Dat Huynh and Ehsan Elhamifar.
\newblock A shared multi-attention framework for multi-label zero-shot
  learning.
\newblock In \emph{Proceedings of the IEEE/CVF conference on computer vision
  and pattern recognition}, pages 8776--8786, 2020.

\bibitem[Kuznetsova et~al.(2020)Kuznetsova, Rom, Alldrin, Uijlings, Krasin,
  Pont-Tuset, Kamali, Popov, Malloci, Kolesnikov, et~al.]{kuznetsova2020open}
Alina Kuznetsova, Hassan Rom, Neil Alldrin, Jasper Uijlings, Ivan Krasin, Jordi
  Pont-Tuset, Shahab Kamali, Stefan Popov, Matteo Malloci, Alexander
  Kolesnikov, et~al.
\newblock The open images dataset v4: Unified image classification, object
  detection, and visual relationship detection at scale.
\newblock \emph{International Journal of Computer Vision}, 128\penalty0
  (7):\penalty0 1956--1981, 2020.

\bibitem[Li et~al.(2014)Li, Zhao, and Guo]{li2014multi}
Xin Li, Feipeng Zhao, and Yuhong Guo.
\newblock Multi-label image classification with a probabilistic label
  enhancement model.
\newblock In \emph{UAI}, pages 1--10, 2014.

\bibitem[Lin et~al.(2014)Lin, Maire, Belongie, Hays, Perona, Ramanan,
  Doll{\'a}r, and Zitnick]{lin2014microsoft}
Tsung-Yi Lin, Michael Maire, Serge Belongie, James Hays, Pietro Perona, Deva
  Ramanan, Piotr Doll{\'a}r, and C~Lawrence Zitnick.
\newblock Microsoft coco: Common objects in context.
\newblock In \emph{European conference on computer vision}, pages 740--755.
  Springer, 2014.

\bibitem[Liu et~al.(2023)Liu, Li, Wu, and Lee]{liu2023visual}
Haotian Liu, Chunyuan Li, Qingyang Wu, and Yong~Jae Lee.
\newblock Visual instruction tuning.
\newblock \emph{arXiv preprint arXiv:2304.08485}, 2023.

\bibitem[Liu et~al.(2021)Liu, Zhang, Yang, Su, and Zhu]{liu2021query2label}
Shilong Liu, Lei Zhang, Xiao Yang, Hang Su, and Jun Zhu.
\newblock Query2label: A simple transformer way to multi-label classification.
\newblock \emph{arXiv preprint arXiv:2107.10834}, 2021.

\bibitem[Loshchilov and Hutter(2017)]{loshchilov2017decoupled}
Ilya Loshchilov and Frank Hutter.
\newblock Decoupled weight decay regularization.
\newblock \emph{arXiv preprint arXiv:1711.05101}, 2017.

\bibitem[Narayan et~al.(2021)Narayan, Gupta, Khan, Khan, Shao, and
  Shah]{narayan2021discriminative}
Sanath Narayan, Akshita Gupta, Salman Khan, Fahad~Shahbaz Khan, Ling Shao, and
  Mubarak Shah.
\newblock Discriminative region-based multi-label zero-shot learning.
\newblock In \emph{Proceedings of the IEEE/CVF International Conference on
  Computer Vision}, pages 8731--8740, 2021.

\bibitem[Pu et~al.(2022)Pu, Chen, Wu, and Lin]{pu2022semantic}
Tao Pu, Tianshui Chen, Hefeng Wu, and Liang Lin.
\newblock Semantic-aware representation blending for multi-label image
  recognition with partial labels.
\newblock In \emph{Proceedings of the AAAI Conference on Artificial
  Intelligence}, pages 2091--2098, 2022.

\bibitem[Radford et~al.(2021)Radford, Kim, Hallacy, Ramesh, Goh, Agarwal,
  Sastry, Askell, Mishkin, Clark, et~al.]{radford2021learning}
Alec Radford, Jong~Wook Kim, Chris Hallacy, Aditya Ramesh, Gabriel Goh,
  Sandhini Agarwal, Girish Sastry, Amanda Askell, Pamela Mishkin, Jack Clark,
  et~al.
\newblock Learning transferable visual models from natural language
  supervision.
\newblock In \emph{International conference on machine learning}, pages
  8748--8763. PMLR, 2021.

\bibitem[Ridnik et~al.(2021)Ridnik, Ben-Baruch, Zamir, Noy, Friedman, Protter,
  and Zelnik-Manor]{ridnik2021asymmetric}
Tal Ridnik, Emanuel Ben-Baruch, Nadav Zamir, Asaf Noy, Itamar Friedman, Matan
  Protter, and Lihi Zelnik-Manor.
\newblock Asymmetric loss for multi-label classification.
\newblock In \emph{Proceedings of the IEEE/CVF International Conference on
  Computer Vision}, pages 82--91, 2021.

\bibitem[Ridnik et~al.(2023)Ridnik, Sharir, Ben-Cohen, Ben-Baruch, and
  Noy]{ridnik2023ml}
Tal Ridnik, Gilad Sharir, Avi Ben-Cohen, Emanuel Ben-Baruch, and Asaf Noy.
\newblock Ml-decoder: Scalable and versatile classification head.
\newblock In \emph{Proceedings of the IEEE/CVF Winter Conference on
  Applications of Computer Vision}, pages 32--41, 2023.

\bibitem[Simon et~al.(2022)Simon, Koniusz, and Harandi]{simon2022meta}
Christian Simon, Piotr Koniusz, and Mehrtash Harandi.
\newblock Meta-learning for multi-label few-shot classification.
\newblock In \emph{Proceedings of the IEEE/CVF winter conference on
  applications of computer vision}, pages 3951--3960, 2022.

\bibitem[Smith(2018)]{smith2018disciplined}
Leslie~N Smith.
\newblock A disciplined approach to neural network hyper-parameters: Part
  1--learning rate, batch size, momentum, and weight decay.
\newblock \emph{arXiv preprint arXiv:1803.09820}, 2018.

\bibitem[Srivastava et~al.(2014)Srivastava, Hinton, Krizhevsky, Sutskever, and
  Salakhutdinov]{srivastava2014dropout}
Nitish Srivastava, Geoffrey Hinton, Alex Krizhevsky, Ilya Sutskever, and Ruslan
  Salakhutdinov.
\newblock Dropout: a simple way to prevent neural networks from overfitting.
\newblock \emph{The journal of machine learning research}, 15\penalty0
  (1):\penalty0 1929--1958, 2014.

\bibitem[Sun et~al.(2022)Sun, Hu, and Saenko]{sun2022dualcoop}
Ximeng Sun, Ping Hu, and Kate Saenko.
\newblock Dualcoop: Fast adaptation to multi-label recognition with limited
  annotations.
\newblock \emph{Advances in Neural Information Processing Systems},
  35:\penalty0 30569--30582, 2022.

\bibitem[Touvron et~al.(2023)Touvron, Lavril, Izacard, Martinet, Lachaux,
  Lacroix, Rozi{\`e}re, Goyal, Hambro, Azhar, et~al.]{touvron2023llama}
Hugo Touvron, Thibaut Lavril, Gautier Izacard, Xavier Martinet, Marie-Anne
  Lachaux, Timoth{\'e}e Lacroix, Baptiste Rozi{\`e}re, Naman Goyal, Eric
  Hambro, Faisal Azhar, et~al.
\newblock Llama: Open and efficient foundation language models.
\newblock \emph{arXiv preprint arXiv:2302.13971}, 2023.

\bibitem[Van~der Maaten and Hinton(2008)]{van2008visualizing}
Laurens Van~der Maaten and Geoffrey Hinton.
\newblock Visualizing data using t-sne.
\newblock \emph{Journal of machine learning research}, 9\penalty0 (11), 2008.

\bibitem[Wang et~al.(2016)Wang, Yang, Mao, Huang, Huang, and Xu]{wang2016cnn}
Jiang Wang, Yi Yang, Junhua Mao, Zhiheng Huang, Chang Huang, and Wei Xu.
\newblock Cnn-rnn: A unified framework for multi-label image classification.
\newblock In \emph{Proceedings of the IEEE conference on computer vision and
  pattern recognition}, pages 2285--2294, 2016.

\bibitem[Wang et~al.(2017)Wang, Chen, Li, Xu, and Lin]{wang2017multi}
Zhouxia Wang, Tianshui Chen, Guanbin Li, Ruijia Xu, and Liang Lin.
\newblock Multi-label image recognition by recurrently discovering attentional
  regions.
\newblock In \emph{Proceedings of the IEEE international conference on computer
  vision}, pages 464--472, 2017.

\bibitem[Wei et~al.(2015)Wei, Xia, Lin, Huang, Ni, Dong, Zhao, and
  Yan]{wei2015hcp}
Yunchao Wei, Wei Xia, Min Lin, Junshi Huang, Bingbing Ni, Jian Dong, Yao Zhao,
  and Shuicheng Yan.
\newblock Hcp: A flexible cnn framework for multi-label image classification.
\newblock \emph{IEEE transactions on pattern analysis and machine
  intelligence}, 38\penalty0 (9):\penalty0 1901--1907, 2015.

\bibitem[Yazici et~al.(2020)Yazici, Gonzalez-Garcia, Ramisa, Twardowski, and
  Weijer]{yazici2020orderless}
Vacit~Oguz Yazici, Abel Gonzalez-Garcia, Arnau Ramisa, Bartlomiej Twardowski,
  and Joost van~de Weijer.
\newblock Orderless recurrent models for multi-label classification.
\newblock In \emph{Proceedings of the IEEE/CVF Conference on Computer Vision
  and Pattern Recognition}, pages 13440--13449, 2020.

\bibitem[Ye et~al.(2020)Ye, He, Peng, Wu, and Qiao]{ye2020attention}
Jin Ye, Junjun He, Xiaojiang Peng, Wenhao Wu, and Yu Qiao.
\newblock Attention-driven dynamic graph convolutional network for multi-label
  image recognition.
\newblock In \emph{European Conference on Computer Vision}, pages 649--665.
  Springer, 2020.

\bibitem[You et~al.(2020)You, Guo, Cui, Long, Bao, and Wen]{you2020cross}
Renchun You, Zhiyao Guo, Lei Cui, Xiang Long, Yingze Bao, and Shilei Wen.
\newblock Cross-modality attention with semantic graph embedding for
  multi-label classification.
\newblock In \emph{Proceedings of the AAAI Conference on Artificial
  Intelligence}, pages 12709--12716, 2020.

\bibitem[Zhang et~al.(2023)Zhang, Huang, Ma, Li, Luo, Xie, Qin, Luo, Li, Liu,
  et~al.]{zhang2023recognize}
Youcai Zhang, Xinyu Huang, Jinyu Ma, Zhaoyang Li, Zhaochuan Luo, Yanchun Xie,
  Yuzhuo Qin, Tong Luo, Yaqian Li, Shilong Liu, et~al.
\newblock Recognize anything: A strong image tagging model.
\newblock \emph{arXiv preprint arXiv:2306.03514}, 2023.

\bibitem[Zhou et~al.(2022{\natexlab{a}})Zhou, Yang, Loy, and
  Liu]{zhou2022conditional}
Kaiyang Zhou, Jingkang Yang, Chen~Change Loy, and Ziwei Liu.
\newblock Conditional prompt learning for vision-language models.
\newblock In \emph{Proceedings of the IEEE/CVF Conference on Computer Vision
  and Pattern Recognition}, pages 16816--16825, 2022{\natexlab{a}}.

\bibitem[Zhou et~al.(2022{\natexlab{b}})Zhou, Yang, Loy, and
  Liu]{zhou2022learning}
Kaiyang Zhou, Jingkang Yang, Chen~Change Loy, and Ziwei Liu.
\newblock Learning to prompt for vision-language models.
\newblock \emph{International Journal of Computer Vision}, 130\penalty0
  (9):\penalty0 2337--2348, 2022{\natexlab{b}}.

\bibitem[Zhu et~al.(2023{\natexlab{a}})Zhu, Chen, Shen, Li, and
  Elhoseiny]{zhu2023minigpt}
Deyao Zhu, Jun Chen, Xiaoqian Shen, Xiang Li, and Mohamed Elhoseiny.
\newblock Minigpt-4: Enhancing vision-language understanding with advanced
  large language models.
\newblock \emph{arXiv preprint arXiv:2304.10592}, 2023{\natexlab{a}}.

\bibitem[Zhu and Wu(2021)]{zhu2021residual}
Ke Zhu and Jianxin Wu.
\newblock Residual attention: A simple but effective method for multi-label
  recognition.
\newblock In \emph{Proceedings of the IEEE/CVF International Conference on
  Computer Vision}, pages 184--193, 2021.

\bibitem[Zhu et~al.(2022)Zhu, Cao, Ge, Liu, and Liu]{zhu2022two}
Xuelin Zhu, Jiuxin Cao, Jiawei Ge, Weijia Liu, and Bo Liu.
\newblock Two-stream transformer for multi-label image classification.
\newblock In \emph{Proceedings of the 30th ACM International Conference on
  Multimedia}, pages 3598--3607, 2022.

\bibitem[Zhu et~al.(2023{\natexlab{b}})Zhu, Liu, Liu, Ge, Liu, and
  Cao]{zhu2023scene}
Xuelin Zhu, Jian Liu, Weijia Liu, Jiawei Ge, Bo Liu, and Jiuxin Cao.
\newblock Scene-aware label graph learning for multi-label image
  classification.
\newblock In \emph{Proceedings of the IEEE/CVF International Conference on
  Computer Vision}, pages 1473--1482, 2023{\natexlab{b}}.

\end{thebibliography}
}

\clearpage
\setcounter{page}{1}
\maketitlesupplementary

\begin{figure}
    \centering
    \includegraphics[width=1\linewidth]{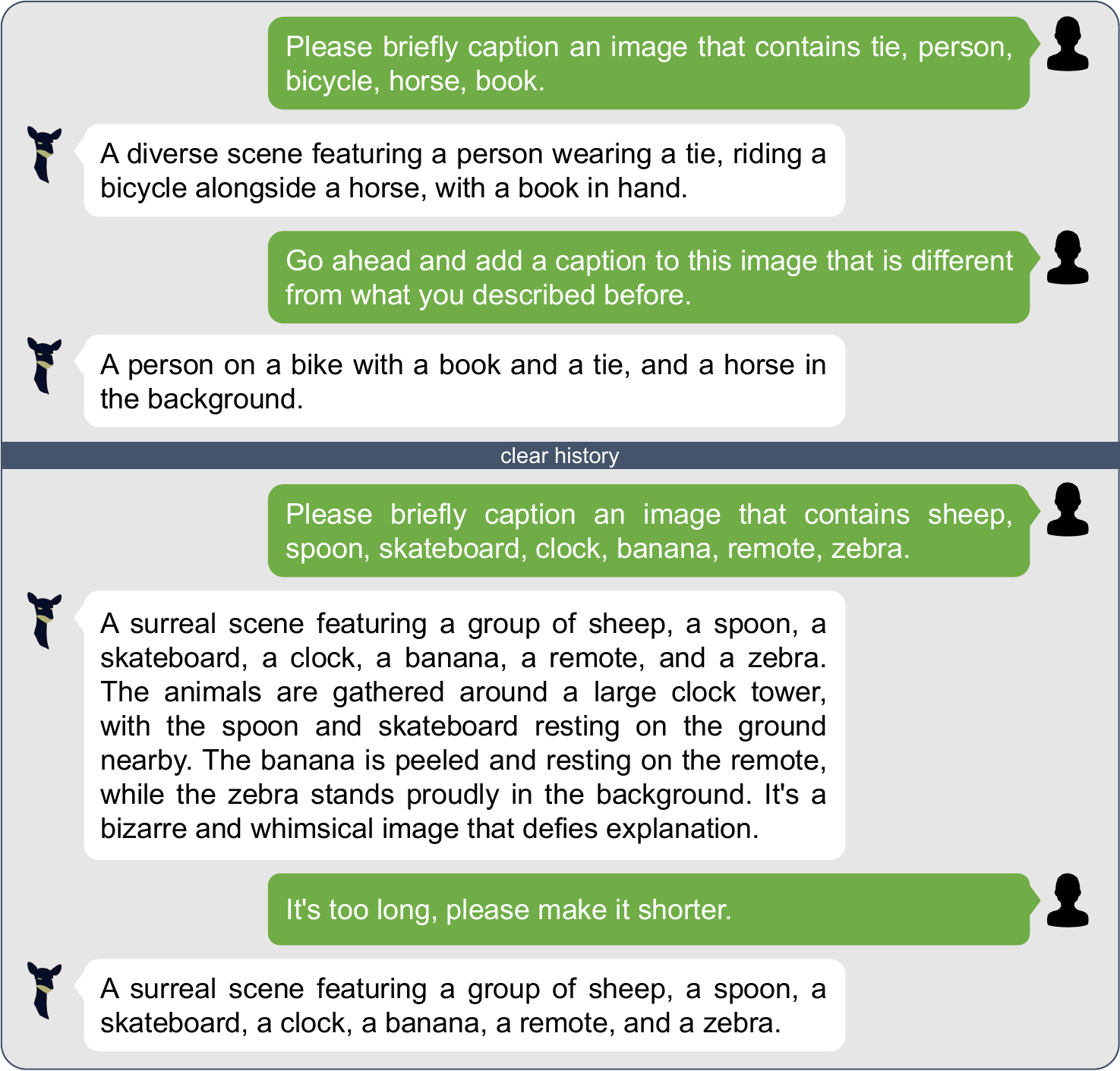}
    \caption{Two examples of using vicuna to generate label-related text in a conversational manner. The upper one shows how to increase the text diversity for each label combination, while the bottom one shows how to get the LLM to generate texts that meet the length requirements.}
    \label{fig:chat}
\end{figure}

\section{Details of LLM-Driven Data Generation}
In order to obtain a large amount of text corpus to train the adapter network, we propose to employ the LLM-based Vicuna \cite{vicuna2023} for multi-label instruction-following text generation in a conversational manner. Details of the conversation are shown in Fig. \ref{fig:chat}. First, we randomly sample several labels and populate them into a well-designed instruction template to drive Vicuna to yield label-related texts. Then, a pre-set instruction, \textit{i.e.}, \textit{Go ahead and add a caption to this image that is different from what you described before}, is designed to generate more diverse texts. Besides, another pre-set instruction (\textit{It's too long, please make it shorter.}) is further used to refine the generated texts to meet the length requirements of the CLIP \cite{radford2021learning} text encoder. With the assistance of these instructions, Vicuna is able to generate a large amount of high-quality text data, thereby ensuring that the adapter network is well trained.

\begin{figure}
    \centering
    \includegraphics[width=1\linewidth]{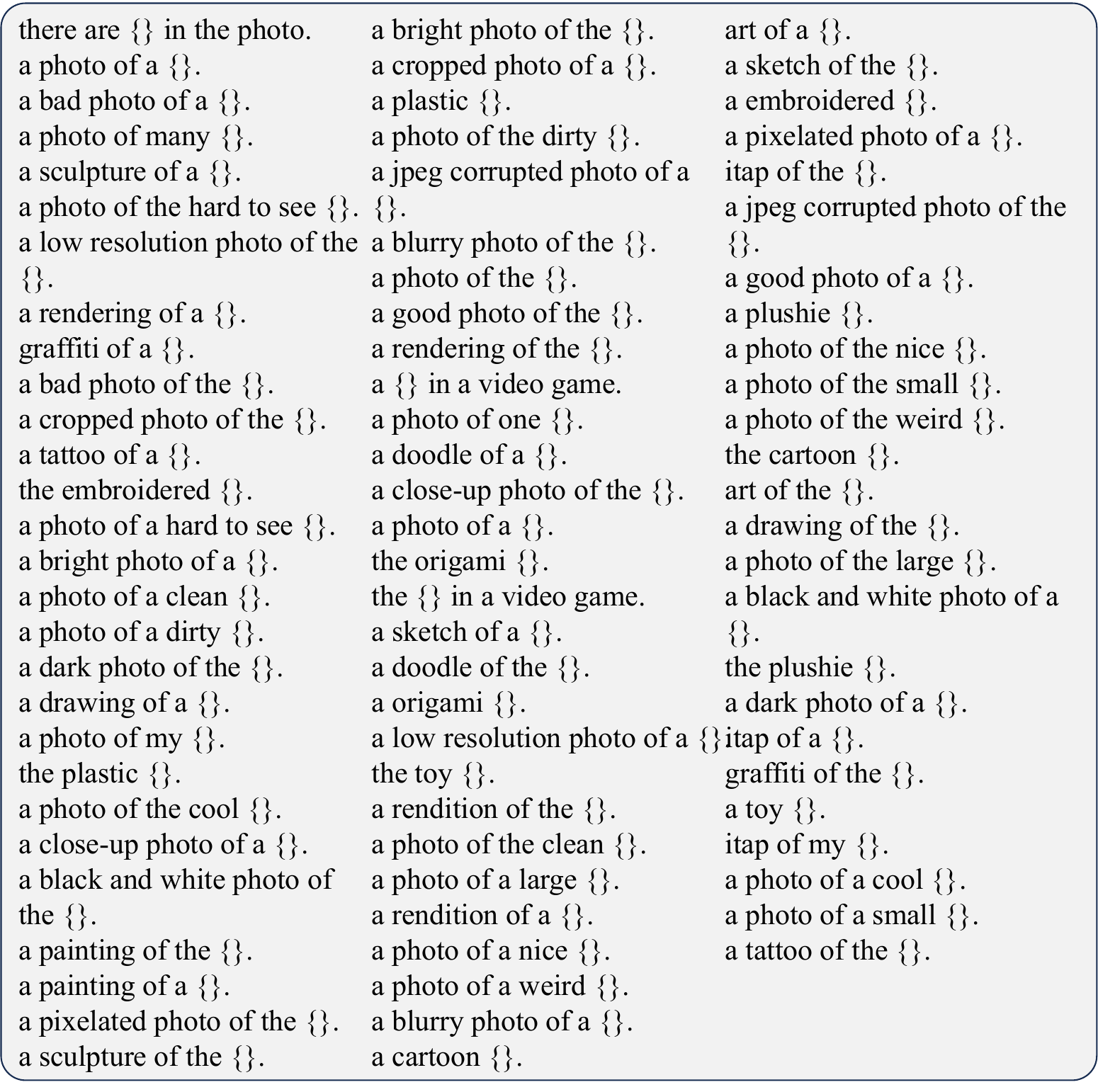}
    \caption{Exhibition of all prompt templates for prompt-based text generation.}
    \label{fig:prompt}
\end{figure}

\section{Details of Prompt-based Text Generation}
In the section 4.3.1, we have discussed the effect of generation method on the performance of the proposed TaI-Adapter framework. In addition to the LLM-based multi-label instruction-following data generation, we also frame a prompt-based data generation method for collecting the desired text corpus. Fig. \ref{fig:prompt} lists all the prompt templates that are able to be directly taken as the training corpus of the adapter network after being populated with randomly sampled labels. Compared to the LLM-based text generation that takes several hours to complete, prompt-based text generation can be done in just a few seconds. This provides a fast solution for visual label recognition. A more interesting idea is to utilize the LLM-based texts and the prompt-based texts simultaneously for the training of our model, which is expected to further improve the performance of our model in multi-label image classification. We leave this as future work to further tap the potential of our method.


\begin{figure*}[htbp]
  \centering
  \begin{subfigure}{0.33\linewidth}
    \includegraphics[width=1.0\linewidth]{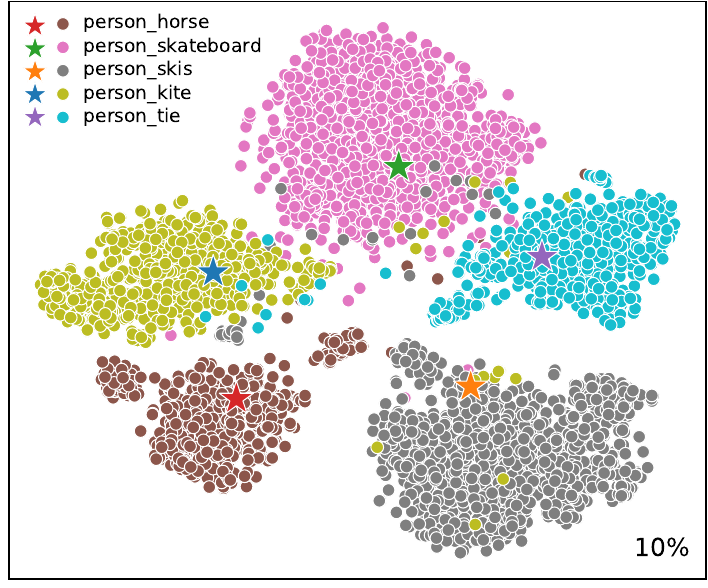}
  \end{subfigure}
  \begin{subfigure}{0.33\linewidth}
    \includegraphics[width=1.0\linewidth]{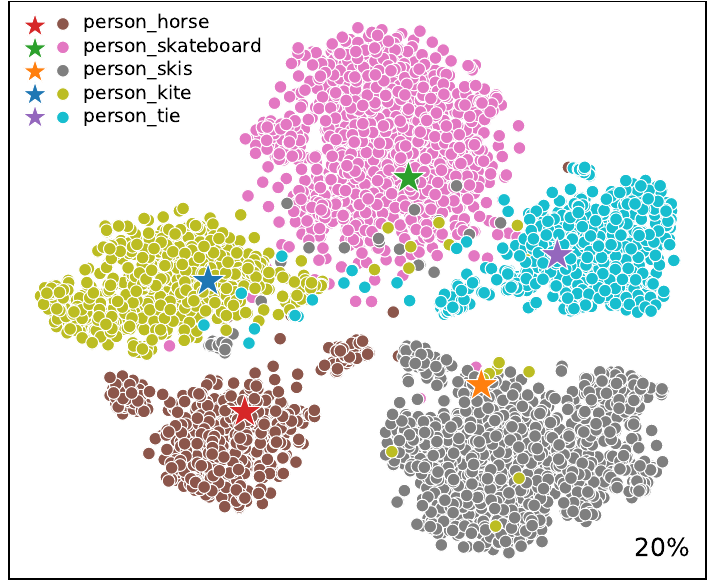}
  \end{subfigure}
  \begin{subfigure}{0.33\linewidth}
    \includegraphics[width=1.0\linewidth]{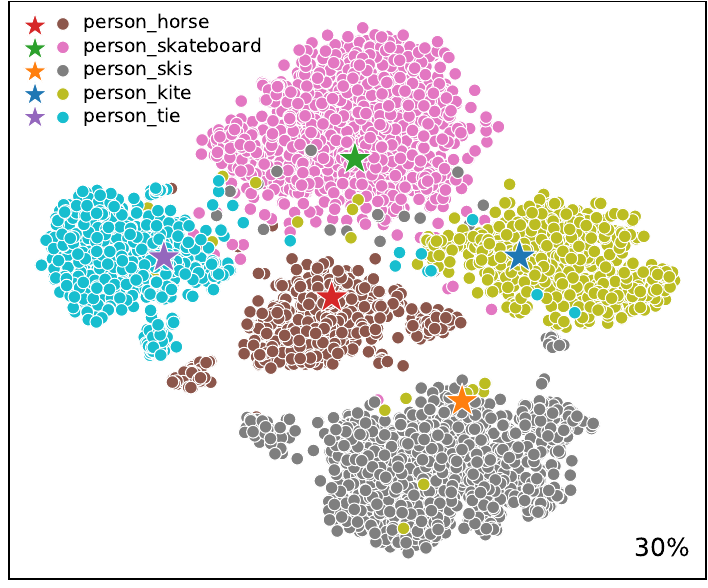}
  \end{subfigure}
  \begin{subfigure}{0.33\linewidth}
    \includegraphics[width=1.0\linewidth]{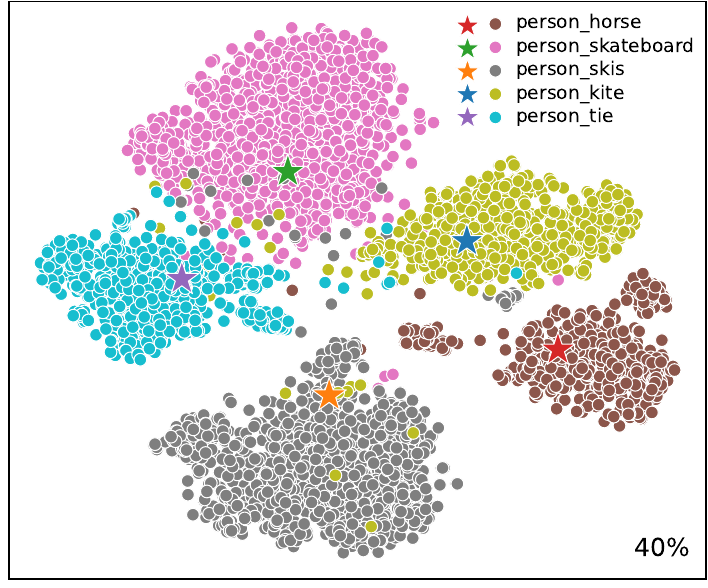}
  \end{subfigure}
  \begin{subfigure}{0.33\linewidth}
    \includegraphics[width=1.0\linewidth]{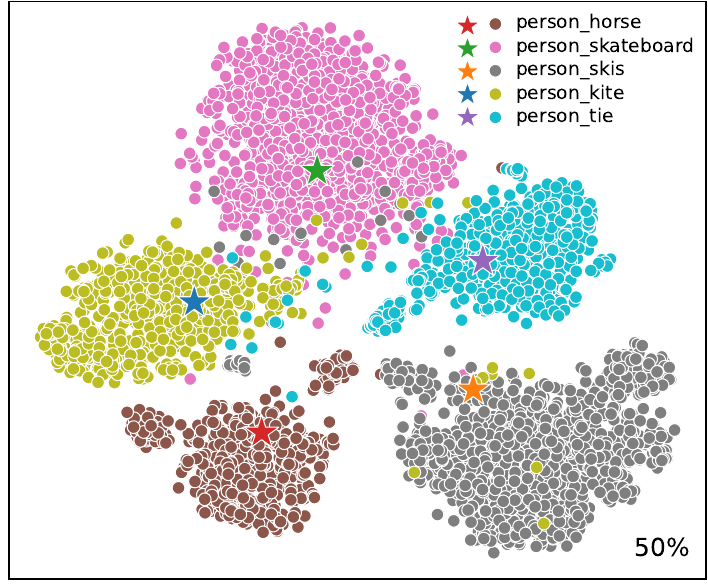}
  \end{subfigure}
  \begin{subfigure}{0.33\linewidth}
    \includegraphics[width=1.0\linewidth]{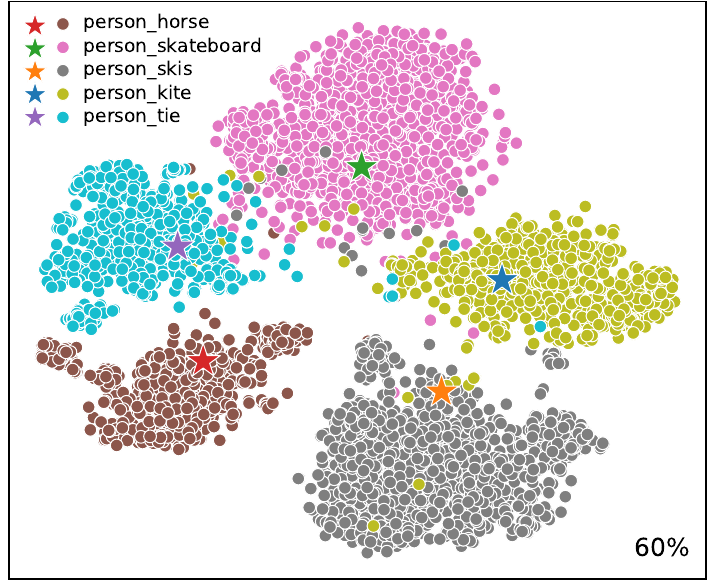}
  \end{subfigure}
  \begin{subfigure}{0.33\linewidth}
    \includegraphics[width=1.0\linewidth]{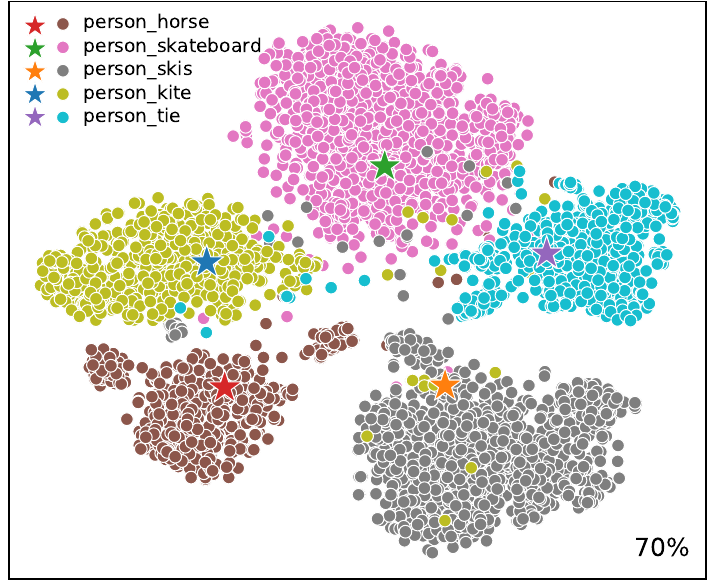}
  \end{subfigure}
  \begin{subfigure}{0.33\linewidth}
    \includegraphics[width=1.0\linewidth]{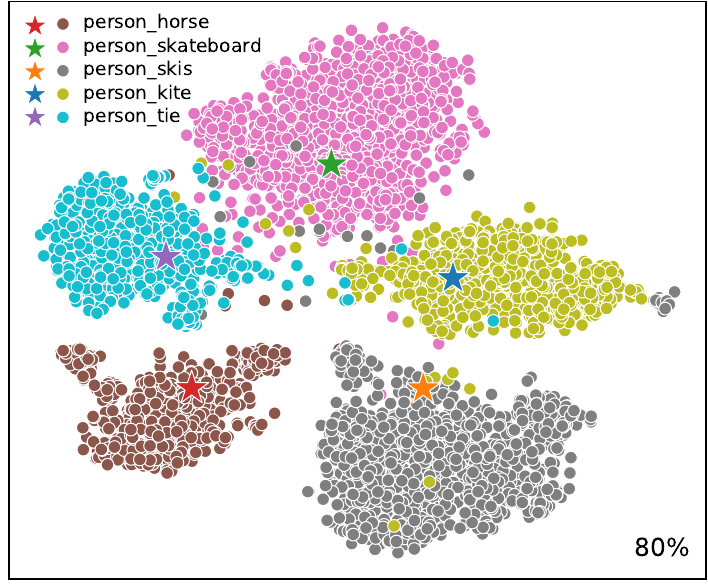}
  \end{subfigure}
  \begin{subfigure}{0.33\linewidth}
    \includegraphics[width=1.0\linewidth]{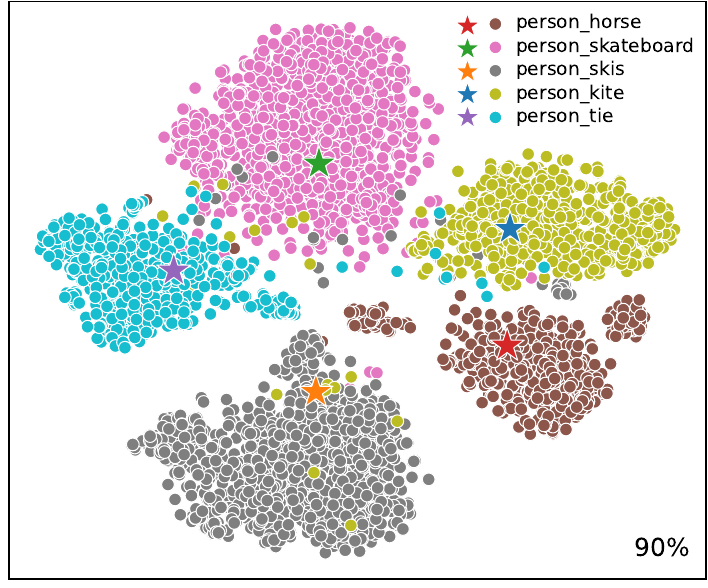}
  \end{subfigure}
  \caption{Visualization on the distribution of image embeddings and estimated centroids (indicated with stars) in the CLIP embedding space using t-SNE \cite{van2008visualizing} for dimensionality reduction. Each figure corresponds to a specific proportion of known labels, identified in the lower right corner.}
  \label{fig:random-PLL}
\end{figure*}

\section{Details of Result Integration}
For a fair comparison with TaI-DPT \cite{guo2023texts} in the multi-label few-shot learning and partial-label learning tasks, we report the model's performance after integrating the results of CoOp \cite{zhou2022learning} and DualCoOp \cite{sun2022dualcoop} in Table 3 and Table 1, respectively. Specifically, we follow the ensemble policy of TaI-DPT and scale the results of CoOp and DualCoOp to the interval from 0 to 1, and then average them with the results of the proposed TaI-Adapter.

\section{More Visualization of Estimated Centroids}
In the section 3.6, to enable the shifted random perturbation mechanism, we propose an insightful method to leverage label-limited image data to estimate the centroids of image embeddings for any label combinations in CLIP embedding space. To further demonstrate its effectiveness and robustness, we choose several label combinations with top image frequency and visualize the estimated centroids as well as the image embeddings in CLIP embedding space. As shown in Fig. \ref{fig:random-PLL}, the centroids estimated from the images with a proportion of labels (from 10\% to 90\%) being known are all aligned with the corresponding image clusters. Despite slightly deviating from the true centroids, these estimated centroids are able to provide an approximate location of image clusters and guide the shifted random perturbation to effectively shift text embeddings closer, allowing the trained adapter to better transfer to the visual modality.

\end{document}